\documentclass[preprints,article,accept,moreauthors,pdftex]{Definitions/mdpi} 

\firstpage{1} 
\pubvolume{1}
\issuenum{1}
\articlenumber{0}
\pubyear{2021}
\copyrightyear{2021}
\datereceived{} 
\dateaccepted{} 
\datepublished{} 
\hreflink{https://doi.org/} % If needed use \linebreak
\usepackage{subcaption}
\Title{Domain-guided Machine Learning for Remotely Sensed In-Season Crop Growth Estimation}

\TitleCitation{Domain-guided Machine Learning for Remotely Sensed In-Season Crop Growth Estimation}

\Author{George Worrall $^{1}$*, Anand Rangarajan $^{2}$ and Jasmeet Judge $^{1}$}

\AuthorNames{George Worrall, Anand Rangarajan and Jasmeet Judge}

\AuthorCitation{Worrall, G.; Rangarajan, A.; Judge, J.}
\address{%
$^{1}$ \quad Center for Remote Sensing, Department of Agricultural and Biological Engineering, University of Florida, Gainesville, FL, USA\\
$^{2}$ \quad Department of Computer \& Information Science \& Engineering, University of Florida, Gainesville, FL, USA}

\corres{Correspondence: gworrall@ufl.edu}

\abstract{Advanced machine learning techniques have been used in remote sensing (RS) applications such as crop mapping and yield prediction, but remain under-utilized for tracking crop progress. In this study, we demonstrate the use of agronomic knowledge of crop growth drivers in a Long Short-Term Memory-based, domain-guided neural network (DgNN) for in-season crop progress estimation. The DgNN uses a branched structure and attention to separate independent crop growth drivers and capture their varying importance throughout the growing season. The DgNN is implemented for corn, using RS data in Iowa for the period 2003-2019, with USDA crop progress reports used as ground truth. State-wide DgNN performance shows significant improvement over sequential and dense-only NN structures, and a widely-used Hidden Markov Model method. The DgNN had a 4.0\% higher Nash-Sutfliffe efficiency over all growth stages and 39\% more weeks with highest cosine similarity than the next best NN during test years. The DgNN and Sequential NN were more robust during periods of abnormal crop progress, though estimating the Silking-Grainfill transition was difficult for all methods. Finally, Uniform Manifold Approximation and Projection visualizations of layer activations showed how LSTM-based NNs separate crop growth time-series differently from a dense-only structure. Results from this study exhibit both the viability of NNs in crop growth stage estimation (CGSE) and the benefits of using domain knowledge. The DgNN methodology presented here can be extended to provide near-real time CGSE of other crops.}


\begin{document}
%%%%%%%%%%%%%%%%%%%%%%%%%%%%%%%%%%%%%%%%%%

\section{Introduction}

The increase in synchrony of global crop production and frequency of climate change-driven abnormal weather events is leading to higher variance in crop yields \cite{ray_climate_2015, iizumi_changes_2016, mehrabi_synchronized_2019}. Most staple food crops are more vulnerable to yield loss in specific stages of growth and as such, accurate crop growth stage estimation (CGSE) is vital to track crop growth at different spatial scales - local, regional, and national - and anticipate and mitigate the effects of variable harvest. High-resolution Remote Sensing (RS) data have been successfully employed to track crop growth at regional scales, however current methods for CGSE utilize curve-fitting and simplistic Machine Learning (ML) models that cannot describe the more complex relationships between crop growth drivers and crop growth stage progress \cite{shen_hidden_2013, zeng_hybrid_2016, seo_improving_2019, diao_remote_2020, ghamghami_parametric_2020}. Many of these methods require full-season data and do not provide in-season CGSE information.

Advanced ML models have found success in applications such as crop-cover mapping/classification and yield estimation \cite{orynbaikyzy_crop_2019, jia_bringing_2019, kerner_resilient_2020, teimouri_novel_2019, Weiss2020}, but these models have yet to been applied for in-season CGSE. Whereas methods such as Neural Networks (NNs) have been used in crop mapping, for which researchers can utilize crop cover maps \cite{service_cropscape_2015} to retrieve millions of crop cover examples per year, field-level crop growth stage (CGS) data is not publicly available and producing field scale ground truth data via field studies is prohibitively expensive. As such, large scale CGSE research relies on local and regional level crop progress data for ground truth. Even for the longest continually running sub-weekly temporal resolution remote sensing (RS) sensors (e.g. MODIS), there are only 21 full growing seasons of crop growth data. Constructing accurate, in-season ML approaches from such limited data is difficult, particularly with few example seasons of abnormal weather. In addition, many crop growth studies estimate events such as `start of season', `peak of season', `end of season' etc. \cite{seo_improving_2019}\cite{diao_remote_2020}, even though these events do not really describe phenological progress and knowledge of their timing may not be actionable. 

Recently, domain knowledge has been used to improve the performance of ML techniques in applied research using techniques collectively known as Theory-guide Machine Learning (TgML) \cite{karpatne_theory-guided_2017}\cite{khandelwal_physics_2020}. TgML techniques include the use of physical models outputs \cite{willard_integrating_2020}, the integration of known domain limits into ML loss functions \cite{karpatne_physics-guided_2018}, and the designing of NN structures that reflect how variables interact within a real physical system \cite{hu_physics-guided_2020}. These techniques have been shown to reduce the amount of data required to reach a given level of performance \cite{karpatne_theory-guided_2017}.  These Theory-guided Neural Networks have begun to significantly improve upon current state-of-the-art methods in applications such as \cite{rong_lagrangian_2020}\cite{wang_deep_2020}. Whereas NNs have shown great promise in agricultural RS studies (e.g. \cite{kerner_resilient_2020}\cite{teimouri_novel_2019}), TgML methods have yet to utilize significant disciplinary advances in agriculture over the last two to three decades. The goal of this study is to understand the impact of incorporating domain knowledge into NN design for in-season CGSE at regional scales. Specifically, the objective of the study is to develop a Domain-guided NN (DgNN) that separates independent growth drivers and compare its performance to sequential NN structures of equivalent complexity. The TgML approach in this study is demonstrated for regional CGSE in field corn, which is one of the most cultivated crops in the world \cite{usda_foreign_agricultural_service_world_2021}. The methodology here, when paired with adequate crop mapping techniques, can be extended to track in-season growth of other crops.

\section{Materials and Methods}
\subsection{Study Area and Data}
\label{sec:study_area}

\begin{figure}[H]
	\centering
	\includegraphics[width=0.5\linewidth]{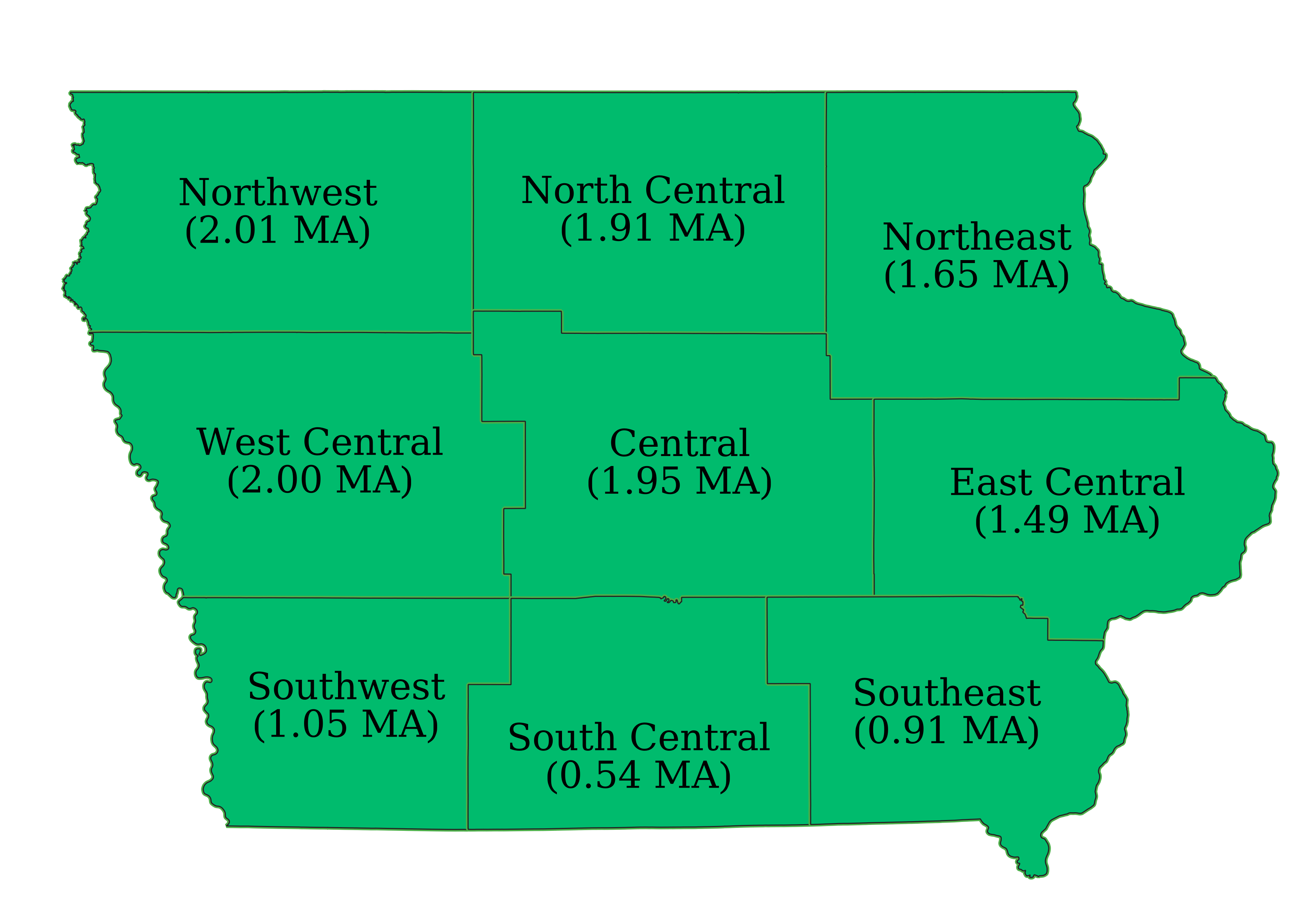}
	\caption{Agricultural Statistical Districts in Iowa with average corn acreage planted over the study period in million acres (MA)  \cite{usda_national_agricultural_statistics_service_corn_nodate}}
	\label{fig:ASDs}
\end{figure}

This study was conducted in the state of Iowa, US, from 2003 to 2019. The state consists of nine separate Agricultural Statistical Districts (ASDs) (see Figure~\ref{fig:ASDs}) and had an average of 13.4 million acres of corn under cultivation across the study period \cite{usda_national_agricultural_statistics_service_corn_nodate}.

\begin{figure}[H]
	\centering
	\includegraphics[width=0.55\linewidth]{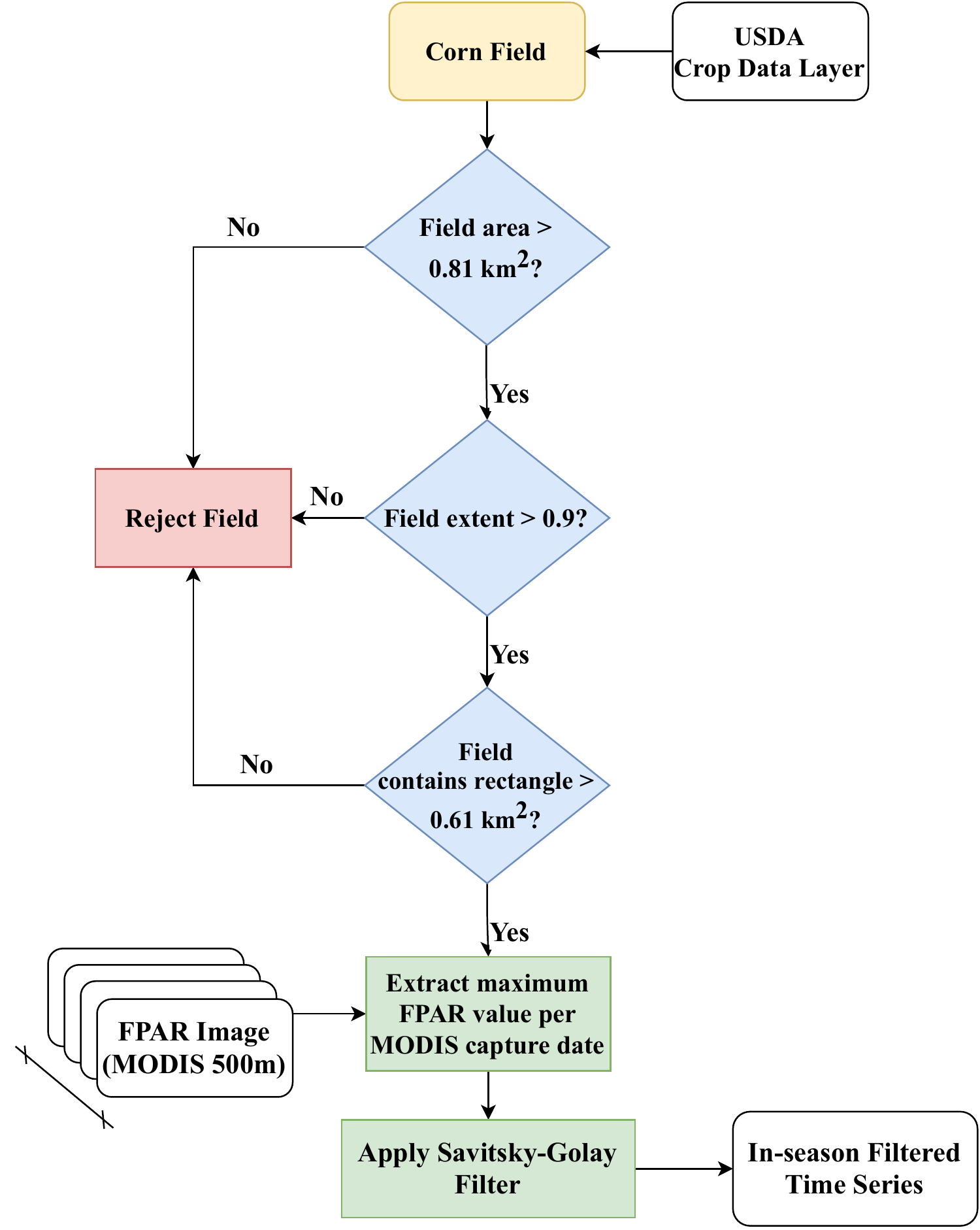}
	\caption{Field selection and processing of MODIS FPAR images.}
	\label{fig:fpar_proc}
\end{figure}

Location of corn field within the study region were obtained from the Corn-Soy Data Layer (CSDL) \cite{wang_mapping_2020} for 2003-2007 and from the USDA Crop Data Layer (CDL) from 2008-2019\cite{usda_ag_data_commons_cropscape_nodate}. In Iowa, corn is typically planted in mid April / early May (week of year (WOY) 15-24), reaches its reproductive stages around late June (WOY 27 onward), and is harvested from early September through late November (WOY 36-48). Weekly USDA-NASS Crop Progress Reports (CPRs), generated from grower and crop assessor surveys, were used as ground truth. CPR progress stages include Planted, Emerged, Silking, Dough, Dent, Mature, and Harvested. In this study, the Planted stage was replaced with Pre-Emergence, a placeholder progress stage that represents all crop/field states prior to emergence, and the Dough and Dent stages were combined as Grainfill. 
CGSE requires both canopy growth information and meteorological data. This study used ASD-wide means and standard deviations of field-level RS and other data shown in Table~\ref{tab:inputs}. Fields in each ASD were selected from the CSDL and CDL based on size and boundary criteria (see Figure~\ref{fig:fpar_proc}).

Micro-meteorological observations obtained from DayMet were used to compute field-level accumulated growing degree days (AGDD), which is a measure of accumulated temperature required for crop growth. AGDD is used to model progression through different corn growth stages, both in remote sensing studies and in mechanistic models \cite{shen_hidden_2013}\cite{ghamghami_parametric_2020}\cite{lizaso_csm-ixim_2011}. The total number of growing degree days (GDD) for a single 24 hour period is calculated using the function:

\[ 
GDD = \left\{
\begin{array}{ll}
	\cfrac{T_{max} + T_{min}}{2} - T_{base} & T_{min} \geq T_{base} \\
	0 & T_{max} < T_{base}\\
\end{array} 
\right. 
\]

where $T_{max}$ is lower value of daily maximum temperature and $34^\circ C$, $T_{min}$ is the minimum recorded daily temperature, and $T_{base}$ is the minimum temperature above which GDD is accumulated, set to $8^\circ C$ \cite{hanks_maize_2015}. AGDD is a running total of daily GDD values, and in this study it is calculated from April 8th of a given year, which is the date prior to first planting during the study period. Solar radiation inputs were converted from W/m$^2$ to MJ/m$^2$/week using day length taken from DayMet to incorporate photoperiod information. Saturated hydraulic conductivities and bulk densities at field centroids from the CSDL and CDL were combined to obtain ASD-wide means and standard deviations. 

\begin{figure}[]
	\widefigure
	\centering
	%\begin{subfigure}[t]{0.7\textwidth}
	%	\centering
	%	\includegraphics[width=\textwidth]{figs/branched.png}
	%	\caption{Branched Model - trainable parameters: 1,013,510}
	%	\vspace{0.5cm}hb
	%	\label{fig:branched_struct}
	%\end{subfigure}%
	\begin{subfigure}[t]{0.65\textwidth}
		\centering
		\includegraphics[width=\textwidth]{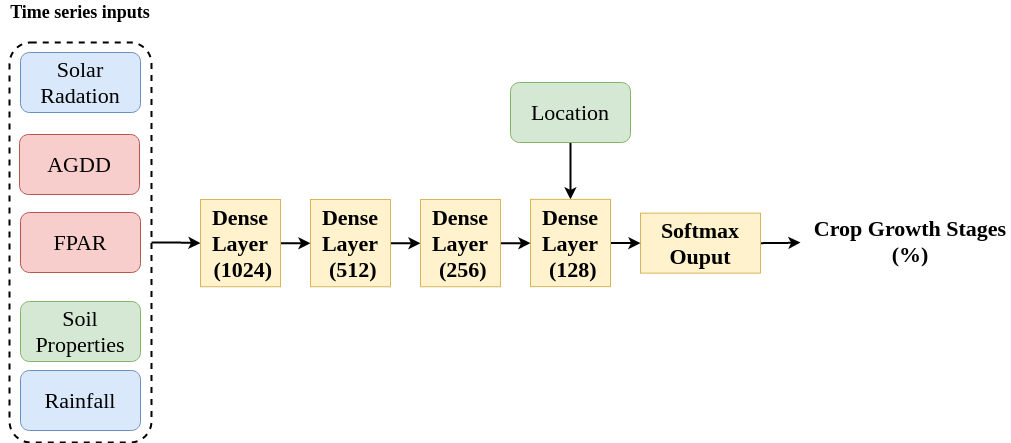}
		\caption{}
		\vspace{0.5cm}
		\label{fig:dense_struct}
	\end{subfigure}
	
	\begin{subfigure}[t]{0.8\textwidth}
		\centering
		\includegraphics[width=\textwidth]{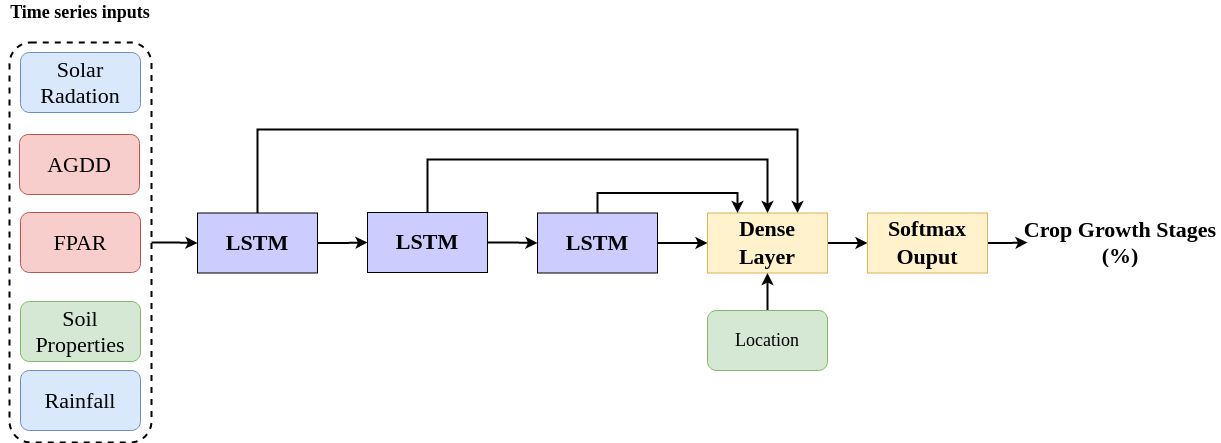}
		\caption{}
		\vspace{0.5cm}
		\label{fig:sequential_struct}
	\end{subfigure}

	\begin{subfigure}[t]{0.7\textwidth}
		\centering
		\includegraphics[width=\textwidth]{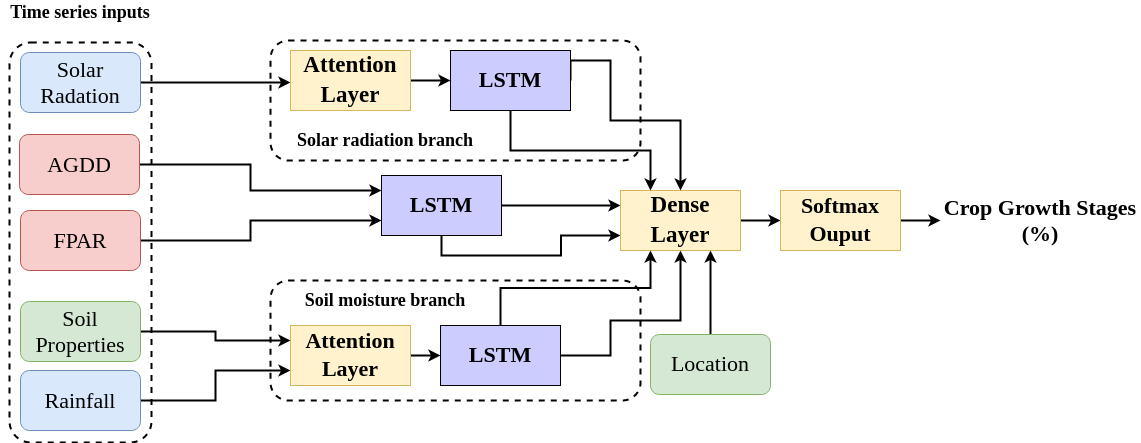}
		\caption{}
		\vspace{0.5cm}
		\label{fig:attention_struct}
	\end{subfigure}
	\caption{Structures of the three NNs implemented in this study. (a) Dense NN, with 1,170,054 trainable parameters. Numbers in parenthesis represent layer node count. (b) Sequential NN, with 1,046,278 trainable parameters. (c) DgNN, with 1,018,094 trainable parameters.}
	\label{fig:structs}
\end{figure}

\end{paracol}
\newpage
\begin{specialtable}[H]
	\widefigure
	\caption{Remote sensing, meteorological, and soil inputs used in this study.}
	\vspace{6pt}
	\label{tab:inputs}
	\scriptsize
	\begin{tabular}{@{}llllcllll@{}}
		\toprule
		Parameter                    & Source                  & Wavelength & Spat. Res. & Temp. Res. & Orig. Units & Input Units      & Reference \\ \midrule
		Solar Radiation              & ORNL DayMet             & -          & 1km        & Daily      & W/m$^2$        & MJ/m$^2$/week & \cite{thornton_daymet_2020}      \\
		Temperature                  & ORNL DayMet             & -          & 1km        & Daily      & degrees C      & AGDD          & \cite{thornton_daymet_2020}      \\
		Rainfall                     & ORNL DayMet             & -          & 1km        & Daily      & mm / day       & mm / week     & \cite{thornton_daymet_2020}      \\
		FPAR                         & NASA MODIS Aqua / Terra & VIS / NIR  & 250m+      & 4-day      & -              & -             & \cite{r_myneni_y_knyazikhin_t_park_mcd15a3h_2015} \\
		Soil Hydrologic Conductivity & USDA-NRSC gSSURGO       & -          & 30m        & Constant   & $\mu$m/s       & $\mu$m/s      & \cite{soil_survey_staff_gridded_2020}          \\
		Soil Bulk Density            & USDA-NRSC gSSURGO       & -          & 30m        & Constant   & g/cm$^3$       & g/cm$^3$      & \cite{soil_survey_staff_gridded_2020}          \\ \bottomrule
	\end{tabular}
\end{specialtable}
\begin{paracol}{2}
\linenumbers
\switchcolumn

To measure canopy growth, 4-day MODIS Fraction of absorbed Photosynthetically Activate Radiation (FPAR) values for each field within an ASD were filtered to produce daily time series data following the Savitsky-Golay (SG) filter method for NDVI used in \cite{chen_simple_2004}. In this study, the SG filter parameters were \textit{m} = 40, \textit{d} = 1 for long-term change trend fitting and m = 4, d = 1 for the main FPAR time series, where 2\textit{m}+1 is the moving filter size and \textit{d} is the degree of the smoothing polynomial. To simulate in-season availability of FPAR data, the values were filtered independently up to each in-season cut-off week. FPAR values for a given week vary slightly as the season progresses and more data is included within the long-term filtering window. Noise filter adaptions to the existing SG method included rejecting points with an absolute gradient of $>$ 0.3 from the previous value, prior to September, the earliest harvest over the study period. These adaptions prevented noisy, phenologically unrealistic data from being included in the moving filter window. It should be noted that while this filtering system is effective for a uni-modal crop such as corn, it may not be effective for crops with more complex seasonal FPAR patterns such as winter wheat, where higher polynomial filter parameters may be required.

\subsection{Data Standardization}
Since CPRs are released every Monday from data collected during the prior week, weekly meteorological data were obtained by aggregating field-scale daily data from Monday-Sunday. The dataset spanned 38 weeks, WOY 13-51, encompassing the earliest planting and latest harvest reported during the study period. To simulate in-season monitoring, one time series was produced from pre-emergence to the `current' week per field, totaling 39 time series. Field-level time series for each input were then aggregated to ASD-level by calculating the mean and standard deviation of the values across each district (median was used for rainfall), with 12 total inputs (Table~\ref{tab:inputs}). These ASD-wide means and standard deviations formed the un-scaled data for in the study. Solar radiation and rainfall were standardized using Z-score scaling and AGDD and FPAR were standardized using MinMax scaling. To standardize the length of each time series to 39 weekly values, all Z-scored in-season time series were zero padded, while MinMax-scaled inputs were padded with 0.5. ASD location within Iowa was represented using a one-hot location vector of length 9, with each bit representing an ASD. The complete 17-year dataset consisted of 5967 time series, each of dimension (39 X 12) with accompanying location vector.

\subsection{Neural Network Design}
The NNs in this study were based upon Long Short-Term Memory (LSTM) layers \cite{hochreiter_long_1997}, that are widely used in sequence identification / classification problems, such as speech recognition, translation, and time series prediction \cite{yu_review_2019}. LSTM has also found success in RS studies, including crop classification \cite{kerner_resilient_2020} and yield prediction \cite{khaki_cnn-rnn_2020}. In this study, LSTM was used because of its ability to handle time series data with variable length gaps between key events \cite{graves_long_2012}, such as variable in-season crop growth data. Three NN implementations were investigated, two reference structures and a third NN that incorporated domain knowledge. The two reference structures included a dense NN, with traditional dense layers of decreasing size (Figure~\ref{fig:dense_struct}) and a sequential NN in which LSTM layers were linearly chained (Figure~\ref{fig:sequential_struct}).

In designing the third NN, interactions between the 12 different inputs (see Table~\ref{tab:inputs}) and their effect on crop growth were considered. Domain knowledge was incorporated into the NN by separating inputs into a branched, structure based on their relationship to crop growth. TgML studies suggest that organizing NN inputs to reflect their real world interactions may improve performance \cite{karpatne_theory-guided_2017}. For example, Khandelwal et al. \cite{khandelwal_physics_2020} were able to improve overall streamflow prediction by 17\% versus traditional LSTM architecture by training dedicated LSTM layers to predict intermediate variables, such as snow pack and soil water content. Although the dataset used in this study does not include target intermediate variables, it is possible to construct a NN structure that encourages LSTM hidden states to learn and track intermediate variables by separating the relevant inputs through a branched structure. For example, excess photoperiod during vegetative stages has been shown to delay crop progress but increase leaf initiation rate \cite{warrington_corn_1983}\cite{warrington_corn_1983-1}, and excess solar radiation or photoperiod is used in mechanistic crop models to determine growth stage timing (e.g. \cite{lizaso_csm-ixim_2011}). In addition, soil moisture stress, due to low rainfall, in juvenile corn has been found to delay growth progression and reduce final plant size \cite{nesmith_short_1992}\cite{cakir_effect_2004}. Typically, the effects of these two drivers on canopy growth and crop progress are modeled separately \cite{lizaso_csm-ixim_2011, jones1983ceres, holzworth_apsim_2014}. In this study, solar radiation and soil moisture-related inputs were separated from FPAR and AGDD using a LSTM-based branched structure, similar to their treatment within agronomic models, and aims to encourage LSTM branches to learn and track these intermediate variables (Figure~\ref{fig:attention_struct}). 

Domain knowledge regarding timing and impact of different crop growth drivers during the growing season was incorporated using an attention mechanism. Attention in NNs allows a network to learn the importance of different inputs. Many natural language processing tasks involving LSTM utilize attention mechanisms to calculate the importance of different words in a sentence, e.g.  \cite{young_recent_2018}\cite{liu_learning_2016}. In this study, self-attention based on Multi-Head Attention \cite{vaswani_attention_2017} is employed to allow the NN to learn importance weightings for different meteorological inputs. Agronomic literature suggests that surplus or deficit of solar radiation and rainfall during particular weeks in the growth cycle are what impact crop growth \cite{warrington_corn_1983}\cite{cakir_effect_2004}. Attention mechanisms were added to both the solar radiation and soil moisture branches to exploit the time-dependent effects of solar radiation and rainfall as growth drivers in corn. The final branched structure with attention mechanisms is shown in Figure~\ref{fig:attention_struct}. This NN is hereafter referred to as the Domain-guided NN (DgNN). 

\end{paracol}
\begin{paracol}{2}
\linenumbers
\switchcolumn

\subsection{NN Loss, Validation and Evaluation}
\label{sec:NN_eval}
In this study, Kullback-Leibler Divergence ($D_{KL}$) was used as the loss function for all NNs. $D_{KL}$ is a measure of the difference between two probability distributions. $D_{KL}$ is often used in NN regression problems with targets that are distributions. Given two distributions $P(x)$a nd $Q(x)$, $D_{KL}$ is calculated as:
\begin{equation}
	D_{KL}(P||Q) = \sum_{x \in X} P(x) \log \left( \frac{P(x)}{Q(x)}\right) 
\end{equation}
Here $D_{KL}$ is used as it provides a measure of the difference between the predicted and actual distributions of crop progress for a given week.

The 17-year CGSE dataset was split into 13 training years and 4 test years, and NN hyperparameters were selected using five-fold cross-validation on the training data. An initial 300 epochs were used in conjunction with early stopping, with a patience of 30 (i.e. 1/10th of the total) epochs and best weights restoration. Early stopping was determined based on NN loss on a single randomly selected year from the training data of each fold, kept separate and assessed at the end of each epoch. NNs were trained using the Adam optimizer with default parameters and a learning rate of 1$e^{-5}$. All NNs stopped early during training. In this study, the Dense NN followed a traditional funnel structure with layer widths ranging from 1024 to 128 nodes. Both the Sequential NN and DgNN used LSTM units with 64 hidden nodes and a 128 node dense layer. Dropout was used between hidden layers with a rate of 0.2. Each NN used a six head softmax output layer, with each head corresponding to one of the six growth stages defined in Section~\ref{sec:study_area}. For self-attention, a two-headed attention layer with a key dimension of 40 was used.

As shown in Tables~\ref{tab:test_years_variables} and~\ref{tab:test_years_stages}, the four test years, 2009, 2012, 2014, and 2019 were chosen based upon deviation from the mean of the dataset in terms of rainfall, planting and harvest dates. All test years remained unseen by NNs during the design phase, and were only used for NN evaluation after selection of the final NN parameters. Initially during evaluation we trained the NNs using the average number of epochs required from early stopping during model validation, but reverted to the same early stopping technique used during NN validation, with the loss on a single randomly selected year (2004) monitored for all NNs.

\begin{specialtable}[H]
	\centering
	\caption{State-wide deviations of precipitation and solar radiation during test years from 2003-2019 study period mean. Standard deviations are given in parenthesis.}
	\label{tab:test_years_variables}
	\scriptsize
	\begin{tabular}{@{}llllll@{}}
		\toprule
		Year & \begin{tabular}[c]{@{}l@{}}AGDD \\ (full season) \\ ($^\circ$C)\end{tabular} & \begin{tabular}[c]{@{}l@{}}Precipitation\\ (April - June) \\ (mm)\end{tabular} & \begin{tabular}[c]{@{}l@{}}Precipitation \\ (full season) \\ (mm)\end{tabular} & \begin{tabular}[c]{@{}l@{}}Solar Radiation \\ (April - June) \\ (W/m$^2$/day)\end{tabular} & \begin{tabular}[c]{@{}l@{}}Solar Radiation \\ (full season) \\ (W/m$^2$/day)\end{tabular} \\ \midrule
		2009 & -277.1 (-2.285)                                              & -52.7 (-0.825)                                                     & +43.7 (+0.388)                                                     & +19.81 (+0.476)                                                                & +5.94 (+0.115)                                                                \\
		2012 & +135.4 (+1.116)                                              & -65.8 (-1.029)                                                     & -193.4 (-1.718)                                                    & +24.00 (+0.576)                                                                & +84.54 (+1.635)                                                               \\
		2014 & -127.0 (-1.047)                                              & +101.2 (+1.584)                                                    & +104.5 (+0.928)                                                    & -20.81 (-0.500)                                                                & -11.94 (-0.231)                                                               \\
		2019 & -80.6 (-0.664)                                               & +15.9 (+0.249)                                                     & +99.9 (+0.887)                                                    & -23.50 (-0.565)                                                                & -87.77 (-1.697)                                                               \\ \bottomrule
	\end{tabular}
\end{specialtable}

% Deviation of growth stages from average of 2003 to 2019
\begin{specialtable}[H]
	\centering
	\caption{State-wide deviations of crop progress during test years from the 2003-2019 study period mean. Values indicate time to cumulative progress into growth stage at 50\%. Standard deviations are given in parenthesis.}
	\label{tab:test_years_stages}
	\scriptsize
	\begin{tabular}{@{}llllll@{}}
		\toprule
		Year & Emerged (days)       & Silking (days)       & Grainfill (days)     & Mature (days)        & Harvested (days)     \\ \midrule
		2009 & -2 (-0.397) & +4 (+0.774) & +11 (+1.58) & +9 (+1.00)  & +20 (+1.98) \\
		2012 & -6 (-1.19)  & -10 (-1.93) & -10 (-1.42) & -16 (-2.01) & -25 (-2.48) \\
		2014 & +3 (+0.60)  & -1 (-0.19)  & -4 (-0.57)  & +6 (+0.75)  & +6 (+0.60)  \\
		2019 & +11 (+2.18) & +2 (+0.39)  & +2 (+0.29)  & +16 (+2.01) & +13 (+1.29) \\ \bottomrule
	\end{tabular}
\end{specialtable}

For comparison to existing CGSE approaches, a Hidden Markov Model-based (HMM) CGSE method presented by Shen et al. was implemented \cite{shen_hidden_2013}. This method uses a standard Expectation Maximization (EM) algorithm along with USDA CPRs to supply priors and transition matrices to the model. Following the method from \cite{shen_hidden_2013}, the HMM was run for 100 runs on the 13 training years, with a 4-year random subset from within the training years selected each run to act as validation data. Average performance over 100 runs was used to reduce EM sensitivity to initialization and local minima. During testing, the HMM was run 10 times on each of the four test years and the mean of these runs is the final performance reported.

ASD-level CGS estimates for the NNs and HMM were aggregated to state-level estimates for comparison via weighted sum, with ASD weights calculated based on the number of corn fields in each ASD that passed the processing criteria, explained in Section~\ref{sec:study_area} and shown in Figure~\ref{fig:fpar_proc}. Performance of the three NN structures and the HMM were evaluated against state-wide USDA CPRs using two metrics. The first, Nash-Sutcliffe efficiency (NSE), is a measure commonly used in hydrology and crop modeling to measure how well a model describes an observed time series versus the mean value of that time series, and is defined as:
\begin{equation}
	NSE = 1 - \frac{\sum_{t=1}^{T}(Q_m^t - Q_o^t)^2}{\sum_{t=1}^{T}(Q_o^t - \bar{Q}_o)^2}
\end{equation}
Where $Q_m^t$ is the model estimate at time $t$, $Q_o^t$ is the observed value at time $t$, and $\bar{Q}_o$ is the mean value of the observed time series. NSE ranges from $-\infty$ to a maximum of 1, where 1 means the model perfectly describes the observed time series. An NSE of below 0 means the model is worse at describing the observed time series than the observed time series mean. In this study, NSE is used as a metric for how well each NN estimates the percentage of corn in a given growth stage over time. 

The second metric, cosine similarity (CS), is a measure of the angle between two vectors in a multi-dimension space. CS between two vectors $\mathbf{A}$ and $\mathbf{B}$ is calculated as:
\begin{equation}
	\textnormal{CS} = cos(\theta) = \frac{\mathbf{A}\cdot\mathbf{B}}{||\mathbf{A}||\ ||\mathbf{B}||}
\end{equation}
In this study, CS is used as a metric for the accuracy of each NN in describing crop progress across all stages for a single week. CS ranges from -1 to +1, representing vectors that are exactly opposite in direction to exactly the same in direction. Orthogonal vectors have a CS of 0. A CS value of 1 means that the CGSE method produces perfect estimates of the amount of corn in each growth stage for that week. Lower CS values indicate higher discrepancies between estimated and real crop progress across all stages for that week.

\subsection{Visualization of NN Operation}
Uniformed Manifold Approximation and Projection (UMAP) \cite{mcinnes_umap_2020} was used to visualize layer activations and gain insight into the differences in behavior among NN structures. UMAP is a dimension reduction technique that is often used for visualizing high-dimension data, e.g. \cite{becht_dimensionality_2019}. For UMAP embedding, local neighborhoods of size 15 were used and hidden layer activations were embedded to 2 dimensions. Layer activations for training and test data were visualized for the layers in each NN feeding into the 128 node dense and softmax layers, these being common to all three NNs (see Figure~\ref{fig:structs}). Color representations of crop progress were formed by reducing the six crop stages to three RGB channels using a UMAP reduction to 3 dimensions, with 15 neighbors.

%%%%%%%%%%%%%%%%%%%%%%%%%%%%%%%%%%%%%%%%%%
\section{Results}

\subsection{Model Validation}
Table~\ref{tab:cross_NSE} shows the state-wide means and standard deviations of NSE for each of the five CGS for five-fold cross-validation on the training data. Overall, the three NN structures performed better than the HMM. The HMM had particularly low NSE during Silking and Mature stages, meaning those stages were difficult to estimate. Figure~\ref{fig:CS_CV_NN} shows five-fold cross-validated CS performance on the training data, with box plots representing the minimum, maximum, and inter-quartile range of the CS. Week-to-week CS for the HMM was also significantly worse than for the NNs. It should be noted that the CS scale for the HMM plot is larger (0.2 to 1) than those used for the NNs in Figure~\ref{fig:CS_CV_NN}.
Among the three NN structures, the Dense NN performed the worst, with the lowest NSE for all stages except Grainfill. In addition, the range of CS values is larger for the Dense NN during transition between Silking and Grainfill (WOY 32-33) and Mature to Harvest (WOY 42-43).

\end{paracol}
\begin{figure}[]
	\centering
	
	%	\begin{subfigure}[t]{\textwidth}
	%		\centering
	%		\includegraphics[width=\textwidth]{figs/branched_cv_base8_CS.png}
	%		\caption{Branched Only}
	%		\vspace{0.2cm}
	%	\end{subfigure}
	\begin{subfigure}[t]{0.9\textwidth}
		\centering
		\caption{}
		\vspace{0.11cm}
		\includegraphics[width=0.9\textwidth]{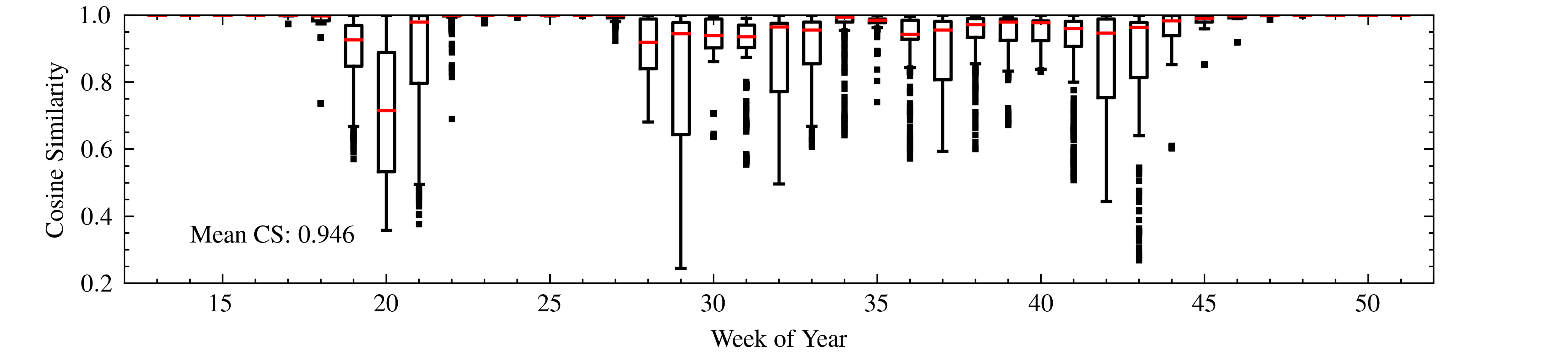}
	\end{subfigure}
	\begin{subfigure}[t]{0.9\textwidth}
		\centering
		\caption{}
		\vspace{0.11cm}
		\includegraphics[width=0.9\textwidth]{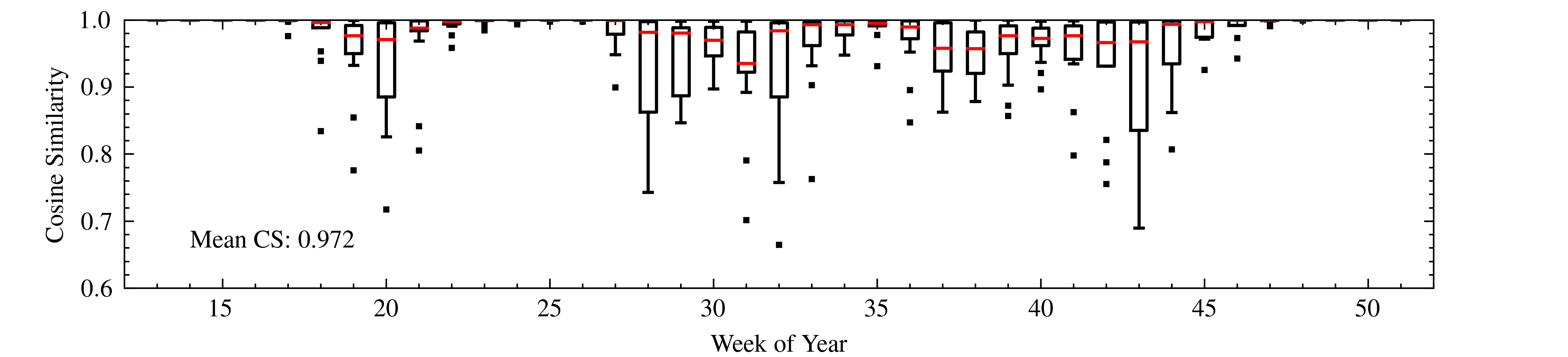}
	\end{subfigure}
	\begin{subfigure}[t]{0.9\textwidth}
		\centering
		\caption{}
		\vspace{0.11cm}
		\includegraphics[width=0.9\textwidth]{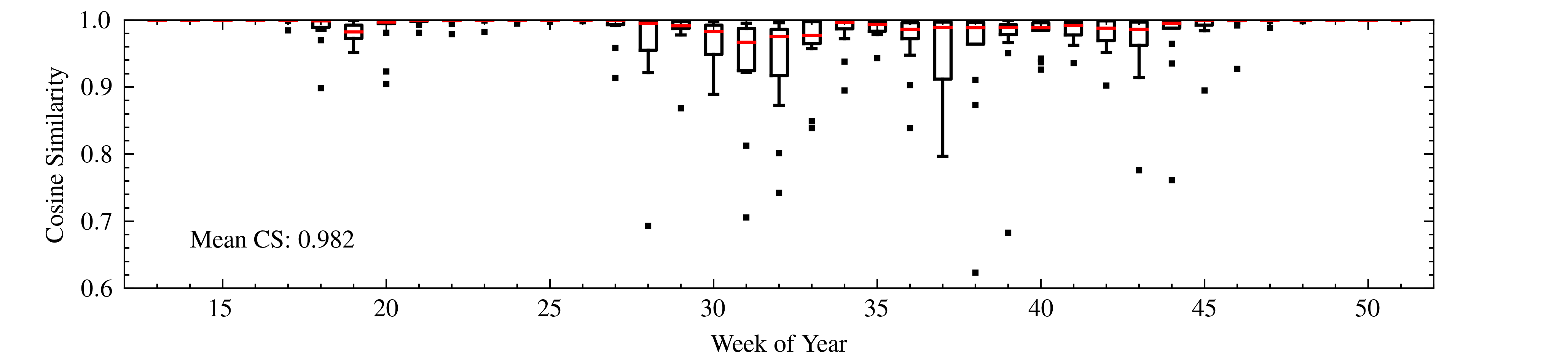}
	\end{subfigure}
	\begin{subfigure}[t]{0.9\textwidth}
		\centering
		\caption{}
		\vspace{0.11cm}
		\includegraphics[width=0.9\textwidth]{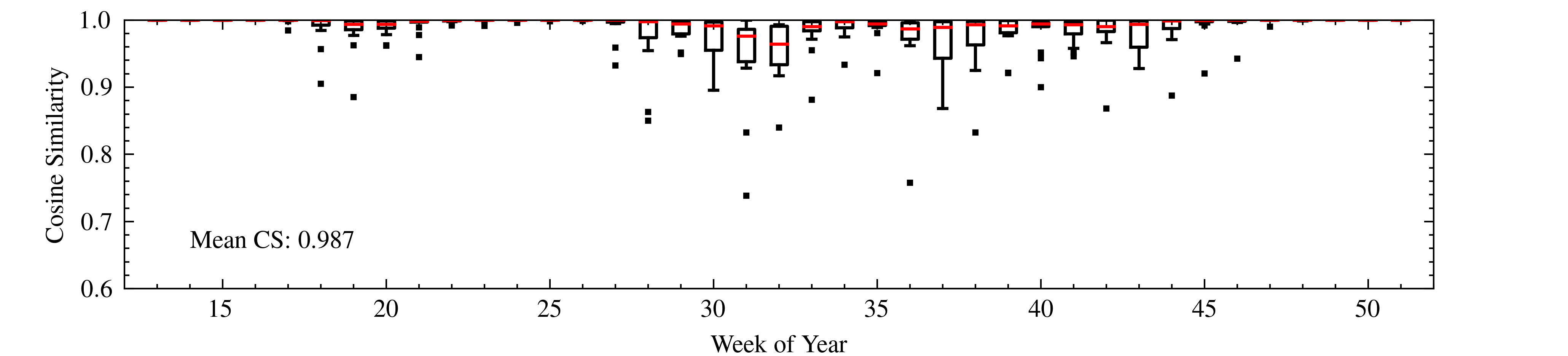}
	\end{subfigure}
	
	\begin{subfigure}[t]{0.91\textwidth}
		\centering
		\caption{}
		\vspace{0.11cm}
		\includegraphics[width=0.91\textwidth]{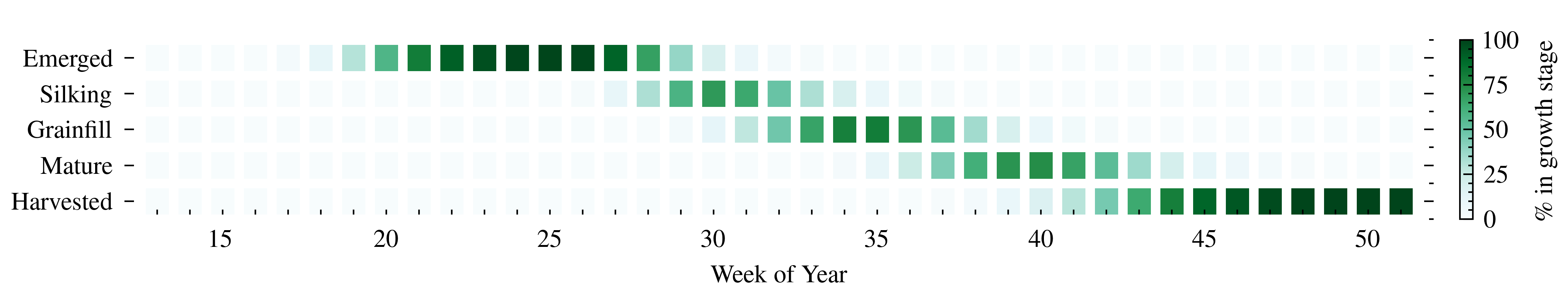}
	\end{subfigure}
	\caption{Week-to-week state-wide CS between real and estimated crop progress of (a) HMM (b) Dense NN (c) Sequential NN (d) DgNN for five-fold cross validation on the 13-year training data. Box plots are mean (red) and inter-quartile range (IQR) and whiskers are 1.5 $\times$ IQR. Outliers are plotted as single points. (e) Average crop progress from USDA crop progress reports during the training years. Scales for HMM (1 to 0.2) and the NNs (1 to 0.6) are different to allow inclusion of all HMM points.}
	\label{fig:CS_CV_NN}
\end{figure}
\begin{paracol}{2}
\linenumbers
\switchcolumn

The Sequential NN improved upon the Dense NN cross-validated NSEs, particularly for the Mature stage. NSE for both the Emerged and Harvested stages for the Sequential NN were also very similar to the DgNN model during validation. Mean CS was only slightly lower for the Sequential NN than for the DgNN, and  the range of CS values over the training data is actually higher during the start of Grainfill (WOY 31). During cross-validation, CGSE proved more challenging for the Silking, Grainfill, and Mature stages. Cross-validated CS performance was high for all methods for weeks 22-27, during the height of the Emerged stage. This is because the Emerged stage is the longest in-season stage and as such, at certain weeks, such as WOY 25, the crop is 100\% emerged for all years during the study period. Similarly, CS performance was high for all methods after WOY 45, after which nearly 100\% of the crop had reached the Harvested stage in all training years.

\begin{specialtable}[H]
	\centering
	\caption{Mean state-wide NSEs and their standard deviations for five-fold cross validation on the 13-year training data. Standard deviations are given in parenthesis. Bold numbers represent highest NSE and lowest standard deviation.}
	\label{tab:cross_NSE}
	\scriptsize
	\begin{tabular}{@{}llllll@{}}
		\toprule	
		Model                                 & NS - Emerged           & NS - Silking           & NS - Grainfill         & NS - Mature            & NS - Harvested         \\ \midrule
		HMM                                   & 0.874 (0.041)          & 0.624 (0.166)         & 0.785 (0.060)           & 0.688 (0.160)          & 0.931 (0.062)  \\
		Dense NN                              & 0.949 (0.041)          & 0.863 (0.123)         & 0.890 (0.062)           & 0.869 (0.087)          & 0.966 (0.034)  \\
		Sequential NN  			              & 0.979 (0.024)	 	   & 0.890 (0.108)         & 0.884 (0.132)           & 0.906 (0.075)          & 0.982 (0.028)          \\
		DgNN                                  & \textbf{0.984 (0.013)} & \textbf{0.914 (0.077)}& \textbf{0.923 (0.063)}  & \textbf{0.935 (0.043)} & \textbf{0.988 (0.014)} \\
		
		% Branched Only NN                      & 0.979 (0.019)          & 0.896 (0.121)         & 0.897 (0.89)            & 0.912 (0.082)          & 0.979 (0.026)  \\
		
		\bottomrule
	\end{tabular}
\end{specialtable}

\subsection{Model Evaluation}
The four test years used for evaluation were 2009, 2012, 2014, and 2019. In 2009, planting was delayed by heavy rains in May and a cool July (6 $^\circ$ C below normal) delayed crop progress. Rain at the end of September and through much of October delayed harvest significantly \cite{usda-national_agricultural_statistics_service_upper_midwest_region_iowa_field_office_2010_2010}. For 2012, corn planting began quickly but was slowed by rain in May. Low rainfall and hot temperatures in June and July caused both soil moisture deficit and fast crop progress. A dry and warm late August and early September brought about early crop maturity and harvesting began early \cite{usda-national_agricultural_statistics_service_upper_midwest_region_iowa_field_office_2013_2013}. Crops progression in 2014 was similar to the 17-year study period average. Planting was slightly behind average during April, but by mid-July corn progression had surpassed the study period average. Wet fields and high grain moisture slowed harvest progress in October \cite{usda-national_agricultural_statistics_service_upper_midwest_region_iowa_field_office_2015_2015}. In 2019, rain in April and May delayed corn planting progress, and by the end of a drier June corn emergence was over a week behind average. Cooler temperatures in September also slowed crop progress, leading to delayed crop maturity and harvesting \cite{usda-national_agricultural_statistics_service_upper_midwest_region_iowa_field_office_2020_2020}.

Table~\ref{tab:test_avg} shows state-wide means and standard deviations of NSE performance for the four test years. The HMM showed lower NSEs during test years for all stages except the Emerged stage. The HMM Mature NSE was significantly lower than in cross-validation, with an average less than 0.3 and standard deviation greater than 0.4. The Dense NN had higher NSEs than the HMM for the four test years, but also a lower mean and higher standard deviation than its cross-validated performance. The Sequential NN produced higher mean and lower standard deviation NSEs than the Dense NN in each stage, and also produced the best NSE for the Silking stage of any method. The DgNN produced the best mean NSE results on the test data for all stages except Silking, with significantly higher performance for the Mature stage. Like all methods, it produced lower mean NSEs than its cross-validated performance on the training data, however it also produced a lower standard deviation for Grainfill performance on the test years.

\begin{specialtable}[H]
	\centering
	\caption{Mean state-wide NSEs and their standard deviations for the four test years. Standard deviations are given in parenthesis. Bold numbers represent highest NSE and lowest standard deviation.}
	\label{tab:test_avg}
	\scriptsize
	\begin{tabular}{@{}llllll@{}}
		\toprule
		Model                                  & NS - Emerged           & NS - Silking           & NS - Grainfill         & NS - Mature            & NS - Harvested         \\ \midrule
		HMM                                    & 0.891 (0.049)          & 0.475 (0.247)          & 0.732 (0.106)          & 0.245 (0.468)          & 0.786 (0.117)          \\
		Dense NN                               & 0.932 (0.016)          & 0.761 (0.094)          & 0.745 (0.185)          & 0.635 (0.375)          & 0.917 (0.062)          \\
		Sequential NN                          & 0.947 (0.048)          & \textbf{0.823 (0.048)} & 0.841 (0.097)          & 0.758 (0.186)          & 0.936 (0.031)           \\
		DgNN                                & \textbf{0.964 (0.015)} & 0.790 (0.127)          & \textbf{0.877 (0.026)} & \textbf{0.870 (0.126)} & \textbf{0.976 (0.021)} \\ \bottomrule
	\end{tabular}
\end{specialtable}

Figures~\ref{fig:2009} through~\ref{fig:2019} show the CS for the four test years. Dense NN CS values for the test years, particularly 2009 (Figure~\ref{fig:2009}) and 2012 (Figure~\ref{fig:2012}), were lower than the range produced during cross-validation. While the Sequential NN was able to produce more accurate overall estimates of all-stage crop progress (as measured by CS) during some of the most difficult periods for all models, it also suffered greater CS performance degradation than the other NNs during some CGS transition periods. The DgNN maintained the highest CS for some of the most difficult periods, such as the Grainfill-Mature-Harvested transitions, during the test years for all NNs. The DgNN was also less prone to CS performance drops during the start and end of the Emergence stage, which were particularly pronounced for the other models in 2009 (WOY 29-30) and 2014 (WOY 18-19). In addition, as seen in Table~\ref{tab:CS_high}, the DgNN produced the best estimates across all growth stages for the greatest number of weeks in each of the test years, producing the highest CS value for 39\% more weeks than the next best NN.

\end{paracol}
\begin{figure}[htbp]
	\centering
	\includegraphics[width=0.9\textwidth]{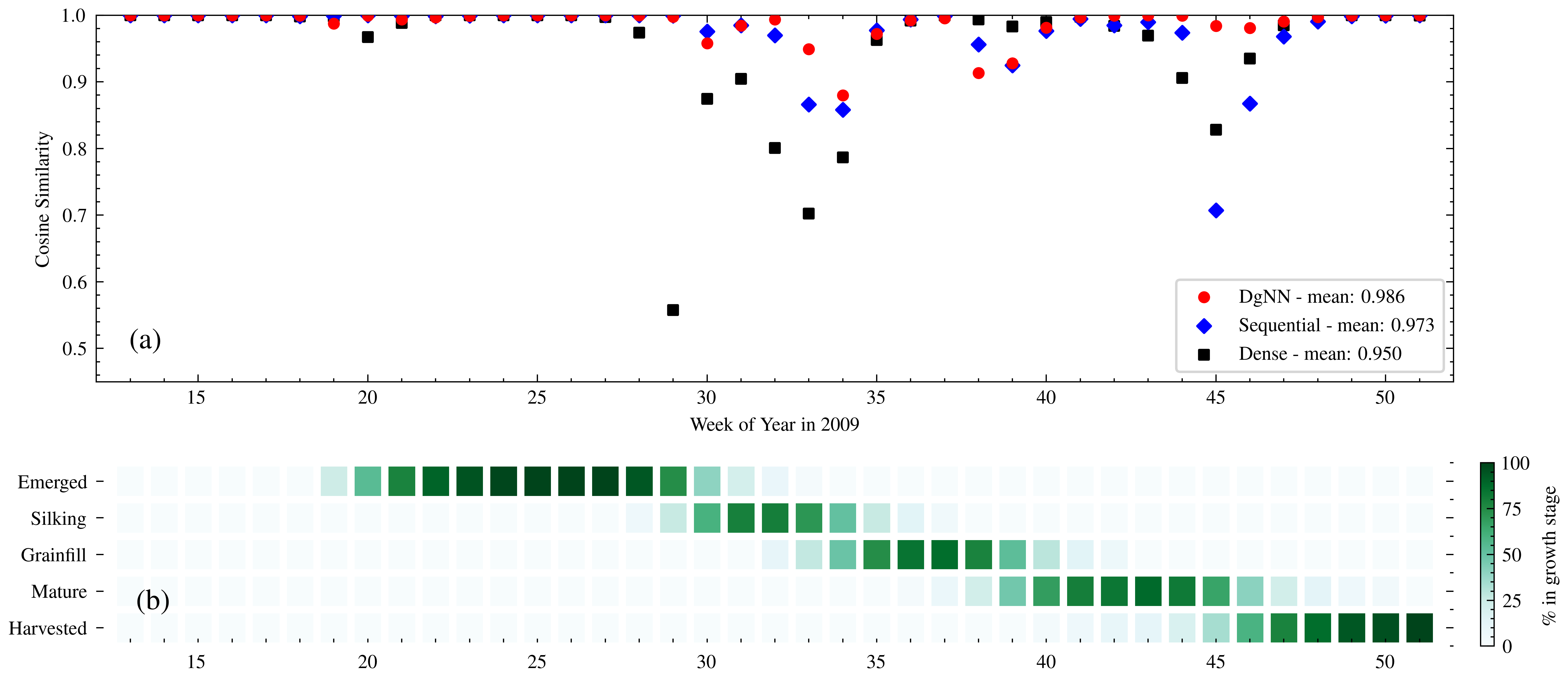}
	\caption{(a) Week-to-week CS between actual and estimated crop progress for the NNs in 2009; and (b) actual crop progress for 2009 is shown below.}
	\label{fig:2009}
\end{figure}

\begin{figure}[htbp]
	\centering
	\includegraphics[width=0.9\textwidth]{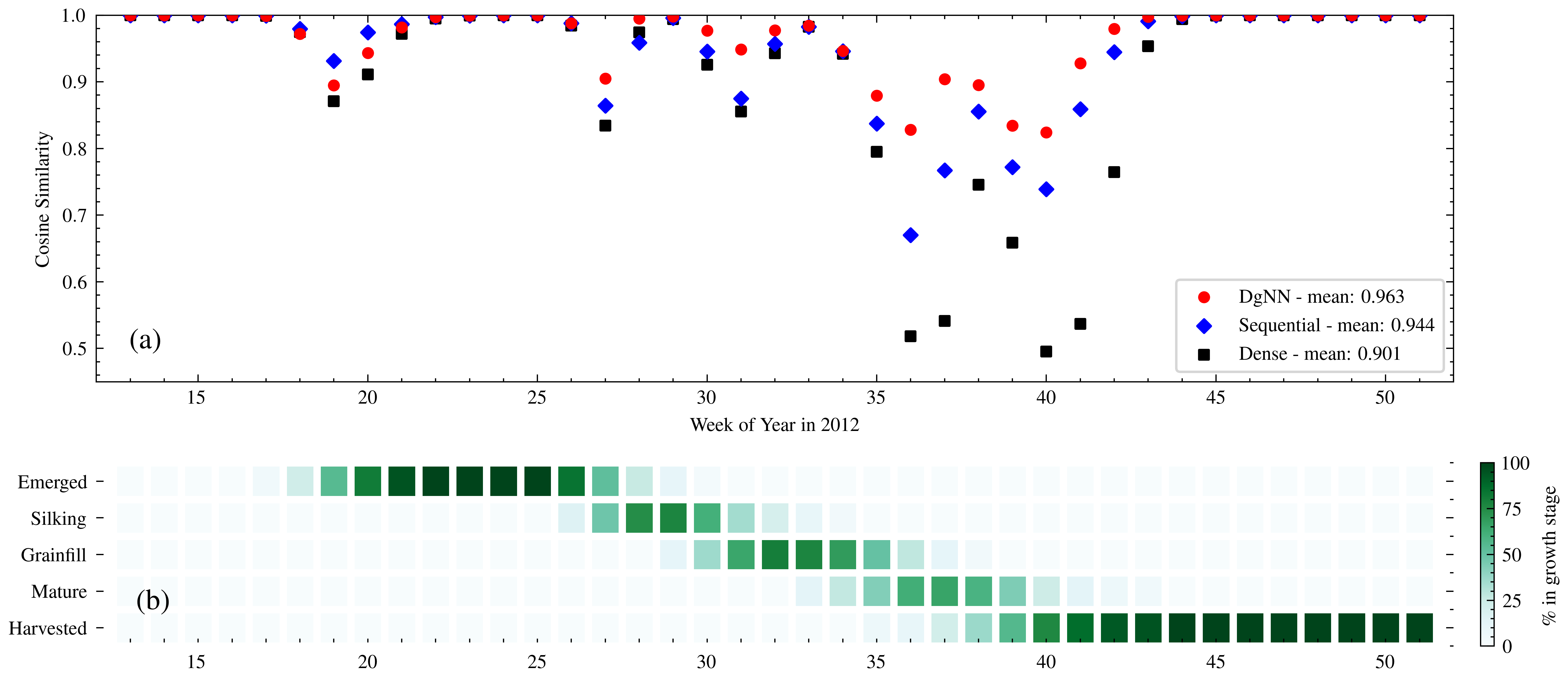}
	\caption{(a) Week-to-week CS between actual and estimated crop progress for the NNs in 2012; and (b) actual crop progress for 2012 is shown below.}
	\label{fig:2012}
\end{figure}

\begin{figure}[htbp]
	\centering
	\includegraphics[width=0.9\textwidth]{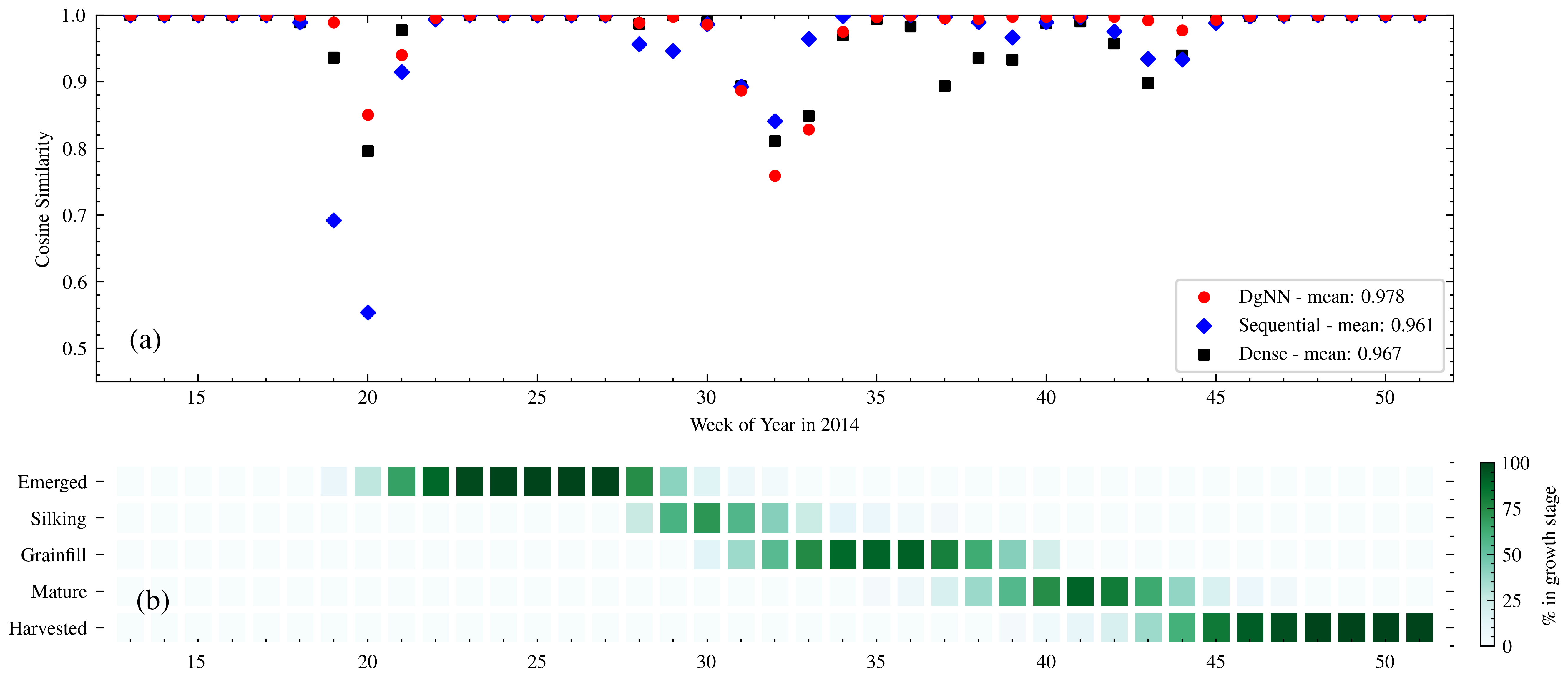}
	\caption{(a) Week-to-week CS between actual and estimated crop progress for the NNs in 2014; and (b) actual crop progress for 2014 is shown below.}
	\label{fig:2014}
\end{figure}

\begin{figure}[htbp]
	\centering
	\includegraphics[width=0.9\textwidth]{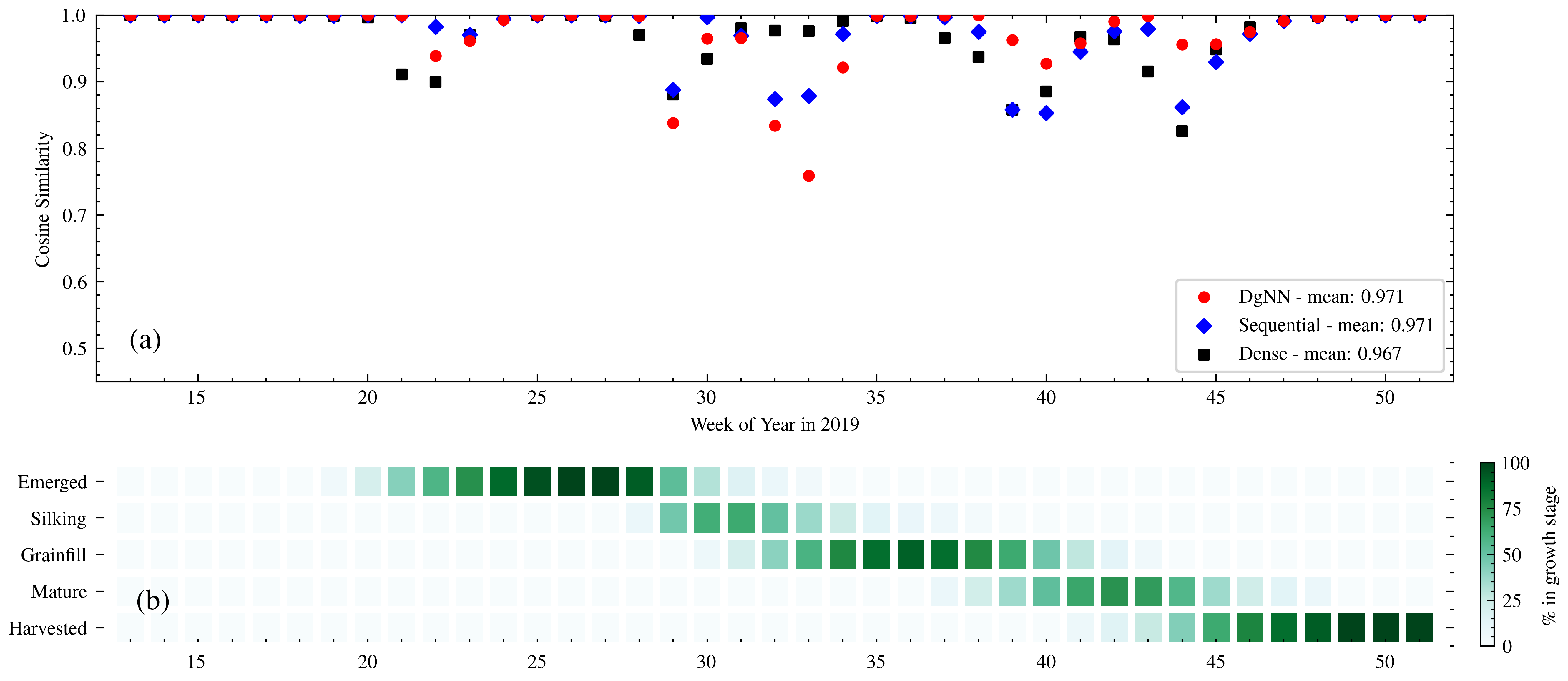}
	\caption{(a) Week-to-week CS between actual and estimated crop progress for the NNs in 2019; and (b) actual crop progress for 2019 is shown below.}
	\label{fig:2019}
\end{figure}
\begin{figure}[htbp]
\centering
\includegraphics[width=\textwidth]{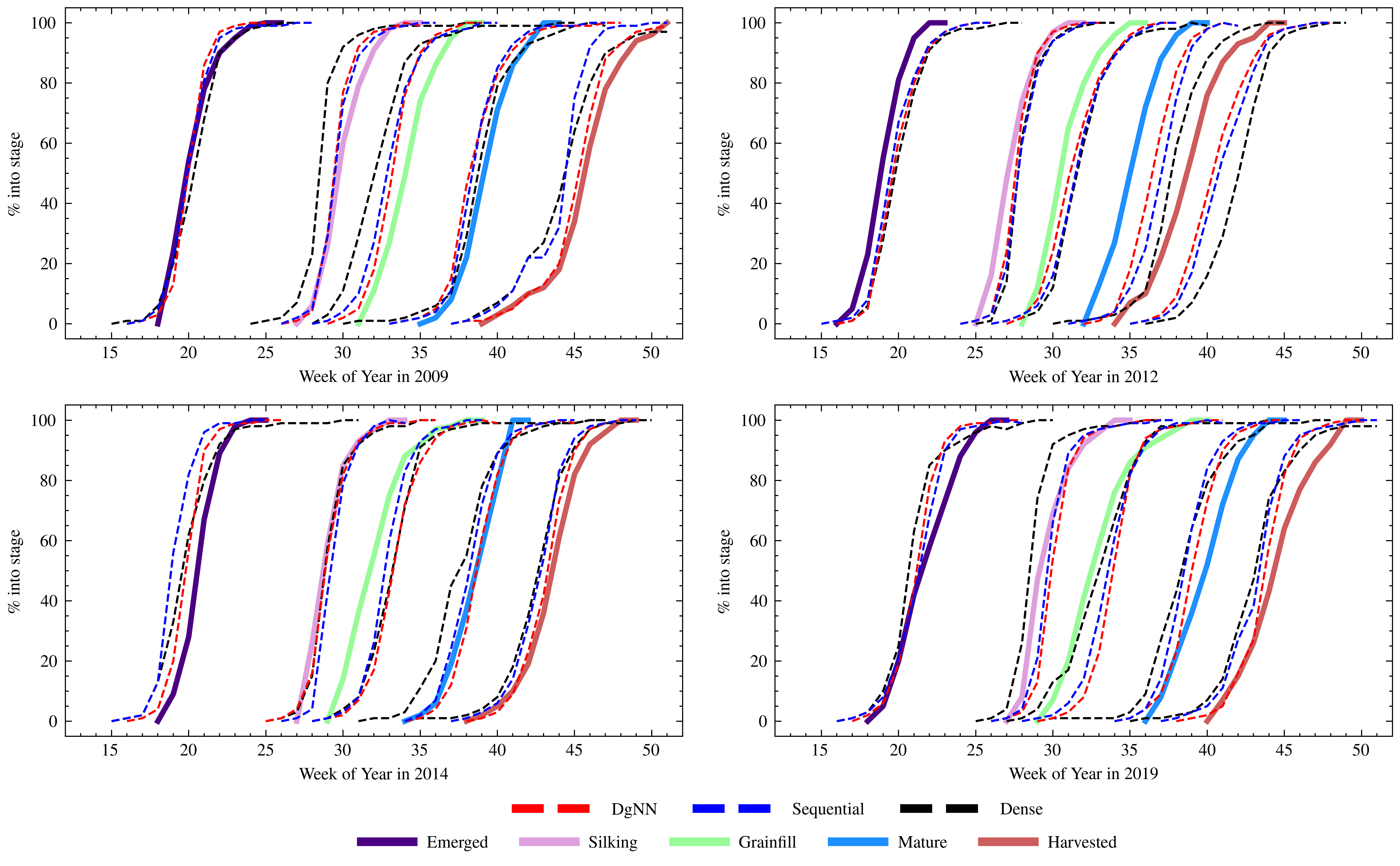}
\caption{State-wide estimated cumulative crop progress of each NN (dashed) for the four test years compared to actual cumulative crop progress (solid).}
\label{fig:cum_test}
\end{figure}

\begin{paracol}{2}
\linenumbers
\switchcolumn

\begin{specialtable}[H]
	\centering
	\caption{Total number of weeks during test years that each NN produced the highest CS value that week. Count includes each NN (one or more) that produced the highest CS value. Bold numbers represent highest number of weeks.}
	\label{tab:CS_high}
	\scriptsize
	\begin{tabular}{@{}llllll@{}}
		\toprule
		Test Year     & 2009        & 2012        & 2014        & 2019        & Total       \\ \midrule
		Dense NN      & 14          & 9           & 14          & 15          & 44          \\
		Sequential NN & 18          & 19          & 18          & 15          & 70          \\
		DgNN          & \textbf{19} & \textbf{32} & \textbf{24} & \textbf{22} & \textbf{97} \\ \bottomrule
	\end{tabular}
\end{specialtable}

Table~\ref{tab:testNSE_years} shows NSE performance for each of the test years. For the 2009 test year, the DgNN outperformed all other models, with the highest NSE for all stages except Emerged. Similar to what was seen during model validation, all NNs had reduced performance during the stages where corn transitioned into and out of Silking (see Figure~\ref{fig:2009}). The Dense NN showed greater performance degradation than the LSTM-based NNs during this period, with significantly lower CS from WOY 29-34. There was a decrease in CS due to a delayed harvest around WOY 45, a time when close to 100\% of the crop had been harvested for every year represented in the training data. The Sequential NN was the worst affected. The DgNN, however, suffered very little decrease in CS during that week. Figure~\ref{fig:cum_test} shows cumulative crop progress for each stage for each of the years. During 2009, each of the three NNs was early in estimating progress for every stage except Emerged.

\begin{specialtable}[H]
	\caption{State-wide NSEs for each of the four test years. Bold numbers represent highest NSE.}
	\label{tab:testNSE_years}
	\centering
	\scriptsize
	\begin{tabular}{@{}llllll@{}}
		\toprule
		Model         & NS - Emerged   & NS - Silking   & NS - Grainfill & NS - Mature    & NS - Harvested \\ \midrule
		\multicolumn{6}{c}{2009}                                                                           \\ \midrule
		HMM           & 0.855          & 0.767          & 0.865          & 0.454          & 0.609          \\
		Dense NN      & 0.915          & 0.610          & 0.781          & 0.904          & 0.933          \\
		Sequential NN & \textbf{0.993} & 0.896          & 0.880          & 0.856          & 0.898          \\
		DgNN          & 0.987          & \textbf{0.917} & \textbf{0.907} & \textbf{0.952} & \textbf{0.991} \\ \midrule
		\multicolumn{6}{c}{2012}                                                                           \\ \midrule
		HMM           & 0.868          & 0.671          & 0.571          & -0.537         & 0.763          \\
		Dense NN      & 0.930          & 0.802          & 0.436          & -0.010          & 0.814          \\
		Sequential NN & \textbf{0.956} & 0.836          & 0.677          & 0.448          & 0.914          \\
		DgNN          & 0.954          & \textbf{0.908} & \textbf{0.837} & \textbf{0.659} & \textbf{0.942} \\ \midrule
		\multicolumn{6}{c}{2014}                                                                           \\ \midrule
		HMM           & \textbf{0.975} & 0.258          & 0.721          & 0.703          & 0.921          \\
		Dense NN      & 0.958          & 0.767 		    & 0.848          & 0.869          & 0.975          \\
		Sequential NN & 0.867          & \textbf{0.789} & \textbf{0.933} & 0.933          & 0.978          \\
		DgNN          & 0.967          & 0.706          & 0.891          & \textbf{0.980} & \textbf{0.994} \\ \midrule
		\multicolumn{6}{c}{2019}                                                                           \\ \midrule
		HMM           & 0.864          & 0.203          & 0.770          & 0.359          & 0.851          \\
		Dense NN      & 0.924          & \textbf{0.865} & \textbf{0.915} & 0.778          & 0.947          \\
		Sequential NN & \textbf{0.973} & 0.771          & 0.873          & 0.797          & 0.954          \\
		DgNN          & 0.948          & 0.627          & 0.875          & \textbf{0.891} & \textbf{0.978} \\ \bottomrule
	\end{tabular}
\end{specialtable}

\end{paracol}
\begin{paracol}{2}
\linenumbers
\switchcolumn

The year 2012 produced the worst NSE across all models for the Grainfill and Mature stages. This is expected given the large deviation in crop progress timing from the study mean during that year (see Table~\ref{tab:test_years_stages}). With the exception of the Emerged stage, the DgNN best described each of the growth stages, as measured by NSE. The week-to-week CS performance degraded less for the DgNN than the other NNs during the fast-moving WOY 33-43 Grainfill-Mature progression, as seen in Figure~\ref{fig:2012}. The pronounced problems of all NNs during this period may have been caused by the weather-induced fast crop progression and sustained drought during the 2012 growing season. Significant soil moisture stress contributed to degradation in corn crop condition \cite{usda_national_agricultural_statistics_service_corn_nodate-1}, which may have affected canopy appearance and the FPAR signal. In addition, warm temperatures during the growing season sped up crop progress to rates not seen since 1987 \cite{usda-national_agricultural_statistics_service_upper_midwest_region_iowa_field_office_2013_2013}. As shown in Table~\ref{tab:test_years_stages}, growth stage time to 50\% was over two standard deviations less the mean for both Mature and Harvested stages. This is in contrast with 2009, where a delay of comparable magnitude (+20 days) did not have as significant an effect on later season week-to-week performance. There were significant delays in NN progress estimation for Mature and Harvested in 2012, as show in Figure~\ref{fig:cum_test}. 

In 2014, Silking NSE for the NNs was low, even though crop progress that year was relative typical of the study period. All three NN structures were late in estimating the onset of the Grainfill stage, as shown in Figure~\ref{fig:cum_test} and also reflected in the decline in CS between WOY 30 and 34 (see Figure~\ref{fig:2014}). Cumulative progress estimates, as seen in Figure~\ref{fig:cum_test}, exhibit the closest cumulative estimation curves for Silking of any of the test years, particularly with the DgNN. However, because the NNs were late in predicting the onset of Grainfill, NSE performance for Silking degraded.  Whereas here the NNs performed well at estimating crop progress into Silking, they could not accurately describe the rate at which the crop progressed out of that stage. NNs were also early in estimating emergence, which degraded CS during WOY 18-21.  

NN performance for Silking also suffered in 2019, with low Silking NSE cause by late DgNN and Sequentiall NN estimates for the start of the Grainfill stage, a problem not experienced by the Dense NN (see Figure~\ref{fig:cum_test}). The opposite is true for the Mature stage, where NSE is higher for the DgNN than the other two NNs. This is reflected in the week-to-week CS values for WOY 37-46 (see Figure~\ref{fig:2019}), higher for the DgNN, during which time the crop was transitioning from Grainfill to Mature to Harvested.

A common theme for all CGSE methods across all training and test years is the three periods of easy estimation that happen at different times during the growing season: WOY 13-17, when crops have yet to emerge, periods around WOY 25 (100\% Emerged for all study years), and WOY 48-51 when the vast majority of the crop has been harvested. Predictably, the most difficult periods for estimation as measured by CS are those with the crop in two or more stages, and progress of mid season stages proved more difficult to estimate than start- and end-of-season stages. Silking is the shortest yet most critical growth stage in terms of potential yield loss, but was the most difficult to estimate for all NNs. Compared to the Emerged and Harvested stages, where significant FPAR gradients help to highlight timing, canopy appearance and therefore FPAR remain relatively unchanged during the Silking-Grainfill transition. Estimation of the this transition was a consistent problem. With the exception of 2009, when the Grainfill stage was delayed by 11 days compared to the study period average, all NNs were late in their estimations of the cumulative progress into the Grainfill stage (see Figure~\ref{fig:cum_test}). In the absence of measurable canopy change, AGDD is a useful proxy for estimating progress from Silking to Grainfill. AGDD, however, is cultivar specific, and cultivars are selected based on different factors, such as planting timing and drought risk. Cultivar-specific variation in required AGDD for progress through Silking and the short duration of that stage may reduce the effectiveness of AGDD as a proxy. In addition, AGDD for this study is accumulated from April 8th of each year and so, due to variable inter-year planting dates, is not an exact measure of how much AGDD that year's crop has accumulated. As the NNs presented here have provided accurate estimates of the timing of progression into Silking, one possible improvement for future methods could be to use weekly GDD as an input so that the NN could learn the accumulation of GDD required relative to the start of Silking. On the other hand, AGDD provides a measure of accumulated temperature over the growing season that serves as an anchor for plausible crop progression. There is no guarantee than NNs presented with GDD only would be able to effectively learn variable accumulation functions that, un-guided, lead to better estimates. 

All NN structures produced NSE and CS results that were worse during model evaluation. This suggests that each of the methods experienced overfitting. In any NN implementation for RS, some overfitting is expected because of the limited number of years of data available for RS-based methods. This leaves CGSE approaches vulnerable to test years with progress timing that is under-represented in the training data. This problem is particularly pronounced in 2012, where fast crop progression saw corn begin to mature in WOY 34, two weeks before any year in the training set. This fast crop progress, however, degraded the performance of the LSTM-based methods less than the HMM and the Dense NN, suggesting that the ability of LSTM to handle variable length gaps between key events make these methods more robust to outlier years such as 2012. Further studies could investigate the relationship between CGSE NN complexity and overfitting through an \textit{a posteriori} ablation study, e.g. \cite{meyes_ablation_2019}.

One notable area of performance degradation were the low NSEs in the DgNN implementation for Silking during model evaluation. In the DgNN structure, self-attention is used to take advantage of field studies in agronomy. In this context, there is one caveat to using attention that may cause overfitting. Since zero-padded, variable length time series were used, the majority of time series inputs in the training-set have later season values set to zero. Therefore, there are less examples of non-zero values during later weeks and the attention layers may tend to reduce the assigned importance of later inputs, introducing a bias. For example, two thirds of all time series in this study have the last one third of their input tensor zero-padded. This could affect the ability of the DgNN to identify important differences later in the season.

Overall, the DgNN showed significant improvement in estimating CGS over the HMM and the other two NNs, particularly in Mature stage NSE and in CS for weeks with multiple overlapping stages. However, the DgNN had reduced NSE for Silking during evaluation due to difficulty in estimating the timing of the Silking-Grainfill transition. Given the higher Silking NSE of the Sequential NN during evaluation, future studies may be able to combine the strengths of both NNs through some form of NN boosting, e.g. \cite{peerlinck_adaboost_2019}.

\subsection{UMAP Visualization of Layer Activations}
Figures~\ref{fig:UMAP_128} and \ref{fig:UMAP_softmax} show layer activations for the final layers of each NN preceding their respective 128 node and softmax layers. Layer activations have been reduced using UMAP and labeled stages in the colorbar represent the height of each new stage. Figure~\ref{fig:UMAP_128} illustrates how the LSTM-based NNs treat the time series differently from the Dense NN. Each of the separate branches visible in the activation space for both the DgNN and Sequential NN contain time series from a specific ASD. The Dense NN, however, does not keep time series from different ASDs as separate, even though it is given the same location information. This may be a source of performance boost for the LSTM-based models, as test year data is projected into the activation space closer to other data from the same ASD. Regression performed by final layers to estimate crop progress is then more strongly influenced by training time series from the same region. This is a benefit as farmers in different ASDs elect to plant cultivars with traits more suited to the local climate. As such, regression performed in an activation space that keeps different ASDs more separate may be more robust to inter-season variation.

The UMAP plots also show evidence of NN overfitting, manifest as noisier delineation among clusters in the test data. For the Dense NN, this is most observable in the Pre-Emergence to Emerged green to purple (\ref{fig:UMAP_softmax} (a) and (b)), and the Mature to Harvested transitions between blue and maroon (\ref{fig:UMAP_128} (a) and (b)). 
While the above is a simple analysis of layer activations using UMAP for model evaluation purposes, UMAP visualizations may be used as a diagnostic tool in future work to assess the impact of including different structures and mechanisms during CGSE NN design.

\end{paracol}
% UMAP plots
\begin{figure}[]
	\centering
	
	\begin{subfigure}{0.45\linewidth}
		\vspace{0.11cm}
		\includegraphics[width=\linewidth]{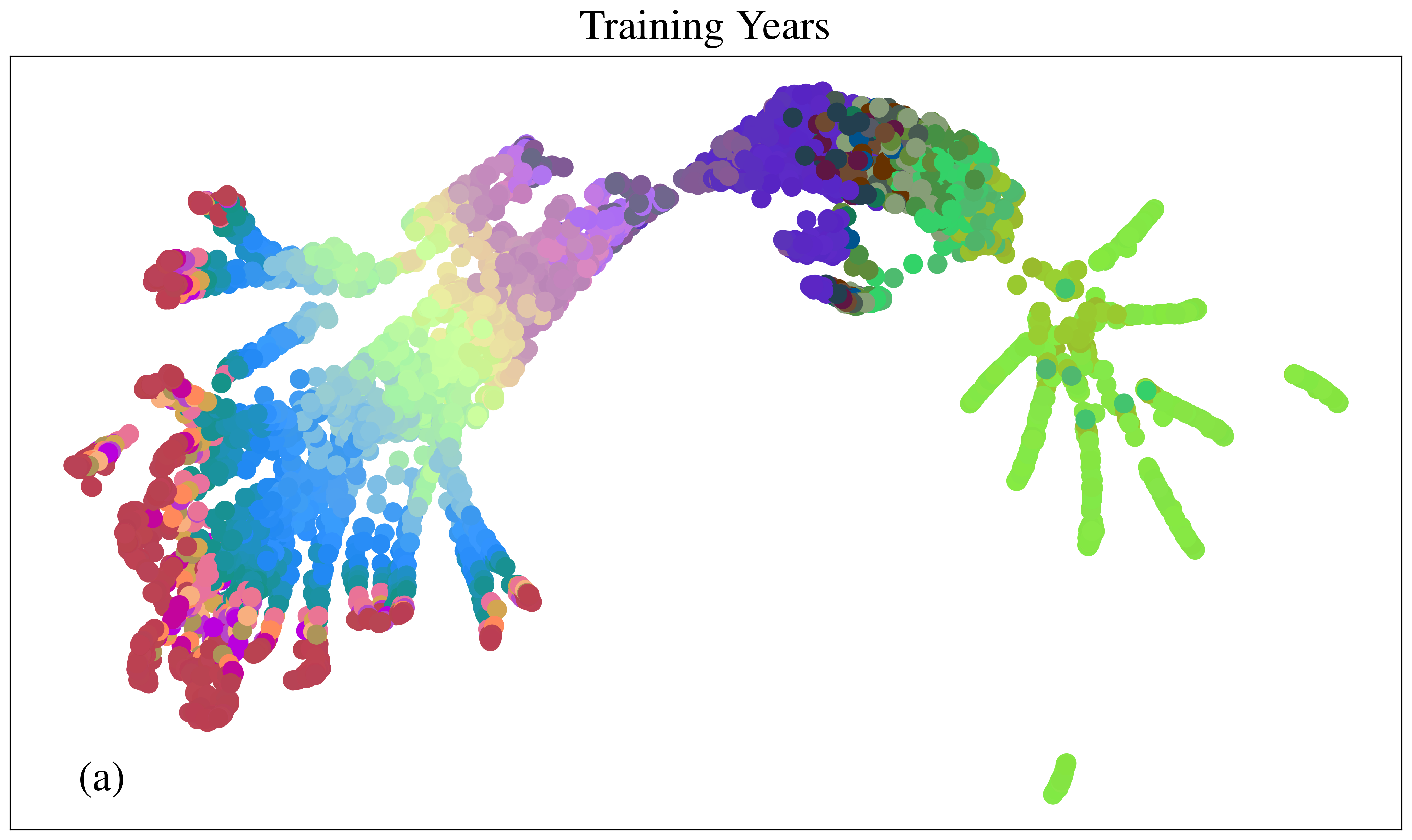}
	\end{subfigure}
	\hfil
	\begin{subfigure}{0.45\linewidth}
		\vspace{0.11cm}
		\includegraphics[width=\linewidth]{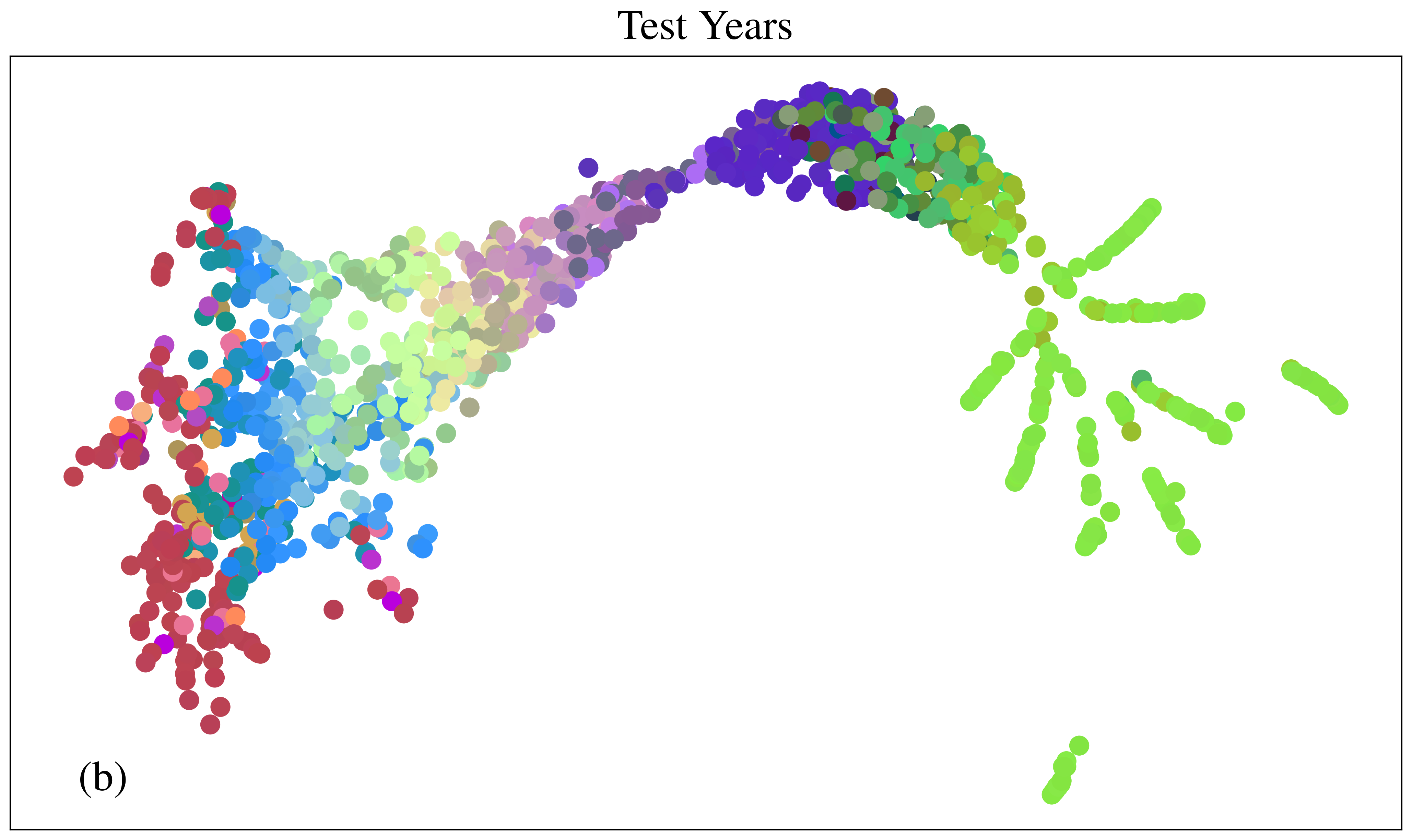}
	\end{subfigure}
	
	\begin{subfigure}{0.45\linewidth}
		\vspace{0.11cm}
		\includegraphics[width=\linewidth]{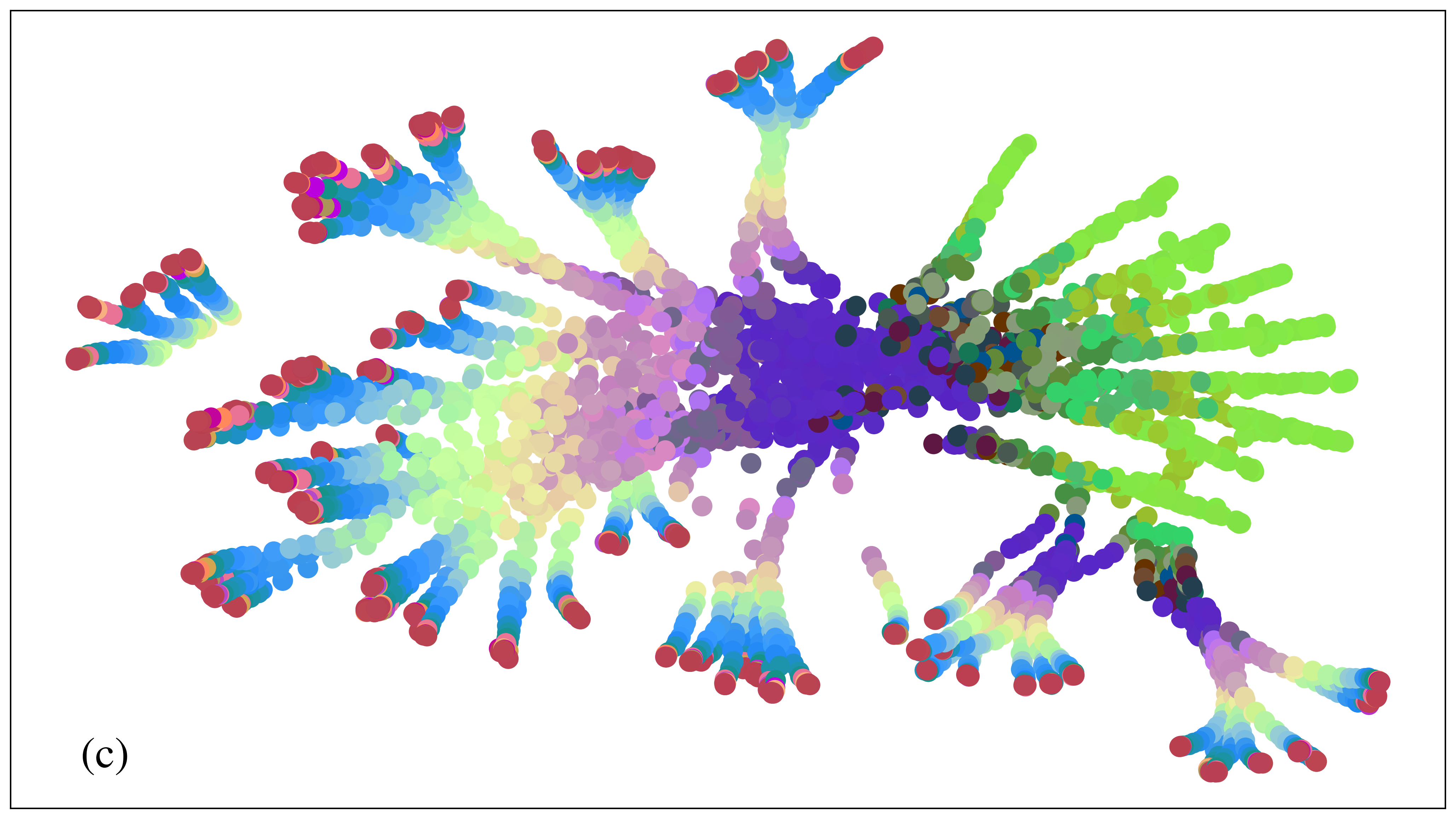}
	\end{subfigure}
	\hfil
	\begin{subfigure}{0.45\linewidth}
		\vspace{0.11cm}
		\includegraphics[width=\linewidth]{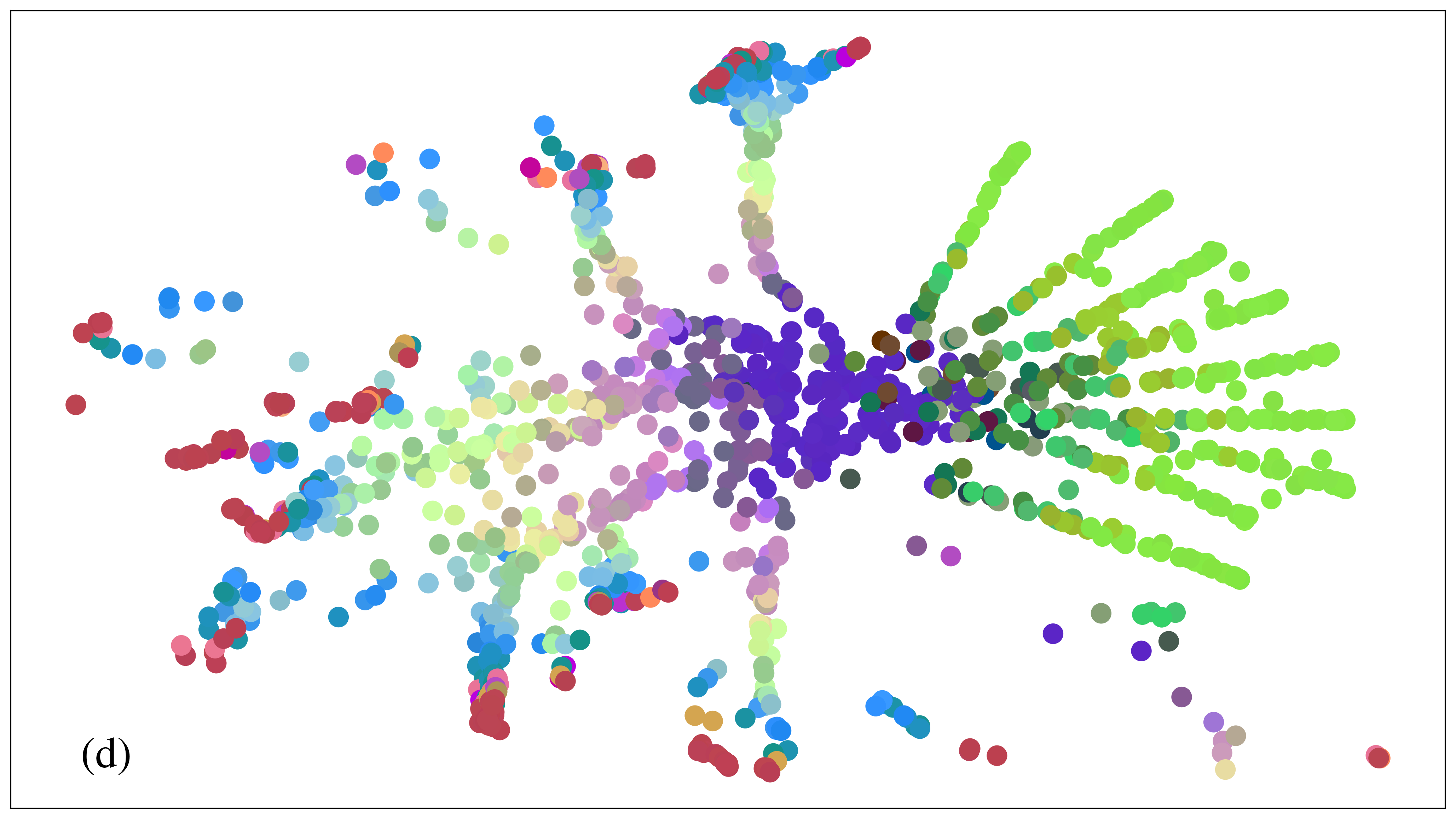}
	\end{subfigure}
	
	\begin{subfigure}{0.45\linewidth}
		\vspace{0.11cm}
		\includegraphics[width=\linewidth]{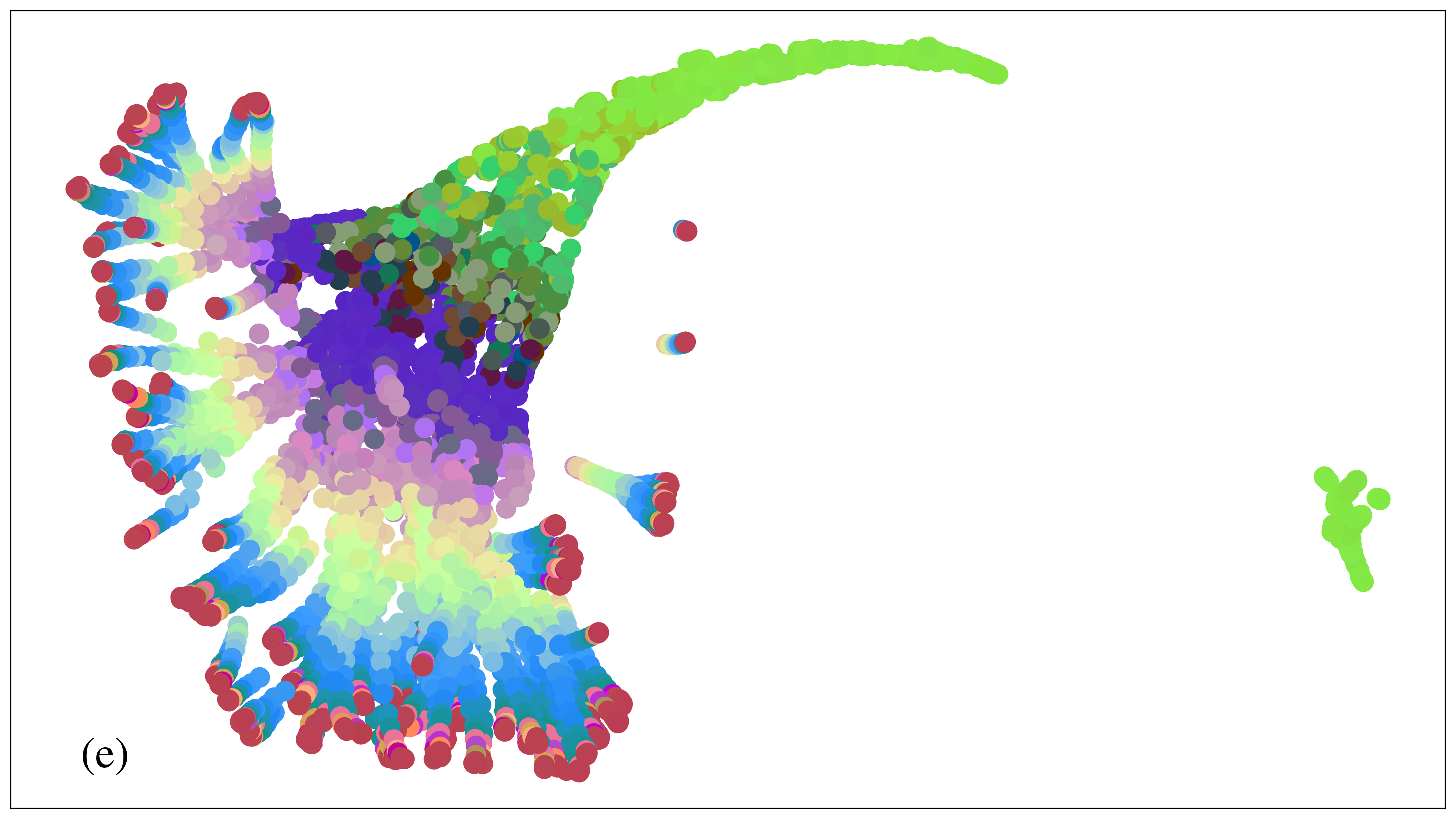}
	\end{subfigure}
	\hfil
	\begin{subfigure}{0.45\linewidth}
		\vspace{0.11cm}
		\includegraphics[width=\linewidth]{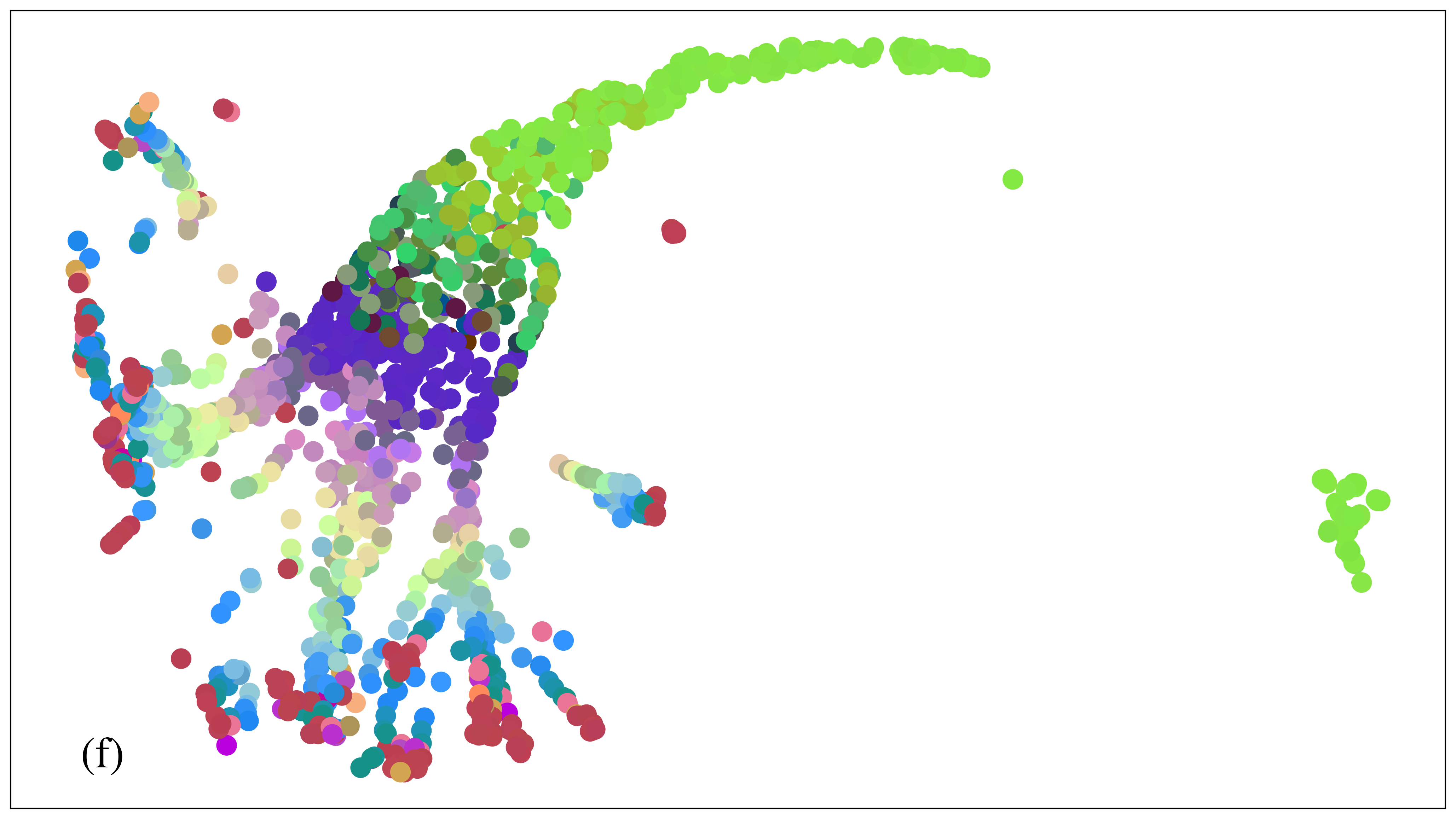}
	\end{subfigure}
	
	\begin{subfigure}{0.9\linewidth}
		\includegraphics[width=\linewidth]{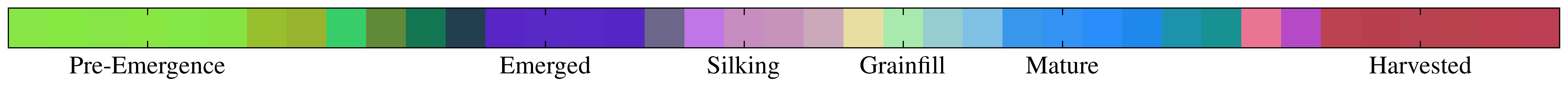}
	\end{subfigure}
	
	\caption{UMAP visualizations of combined activations from layers feeding into the 128 node layer for training and test years for (a) and (b) Dense NN; (c) and (d) Sequential NN; (e) and (f) DgNN.}
	\label{fig:UMAP_128}
\end{figure}

% UMAP plots
\begin{figure}[]
	\centering
	
	\begin{subfigure}{0.45\linewidth}
		\includegraphics[width=\linewidth]{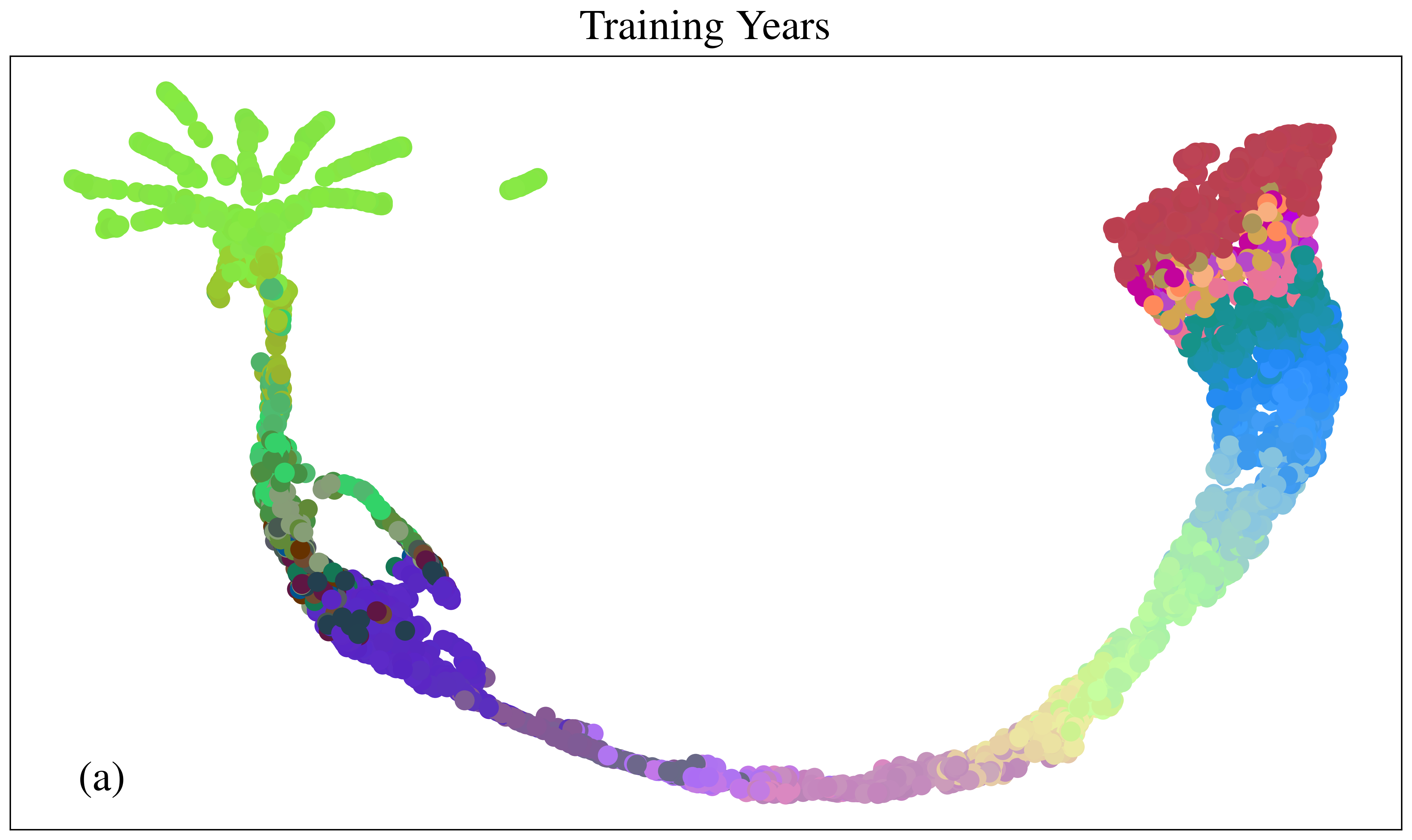}
	\end{subfigure}
	\hfil
	\begin{subfigure}{0.45\linewidth}
		\includegraphics[width=\linewidth]{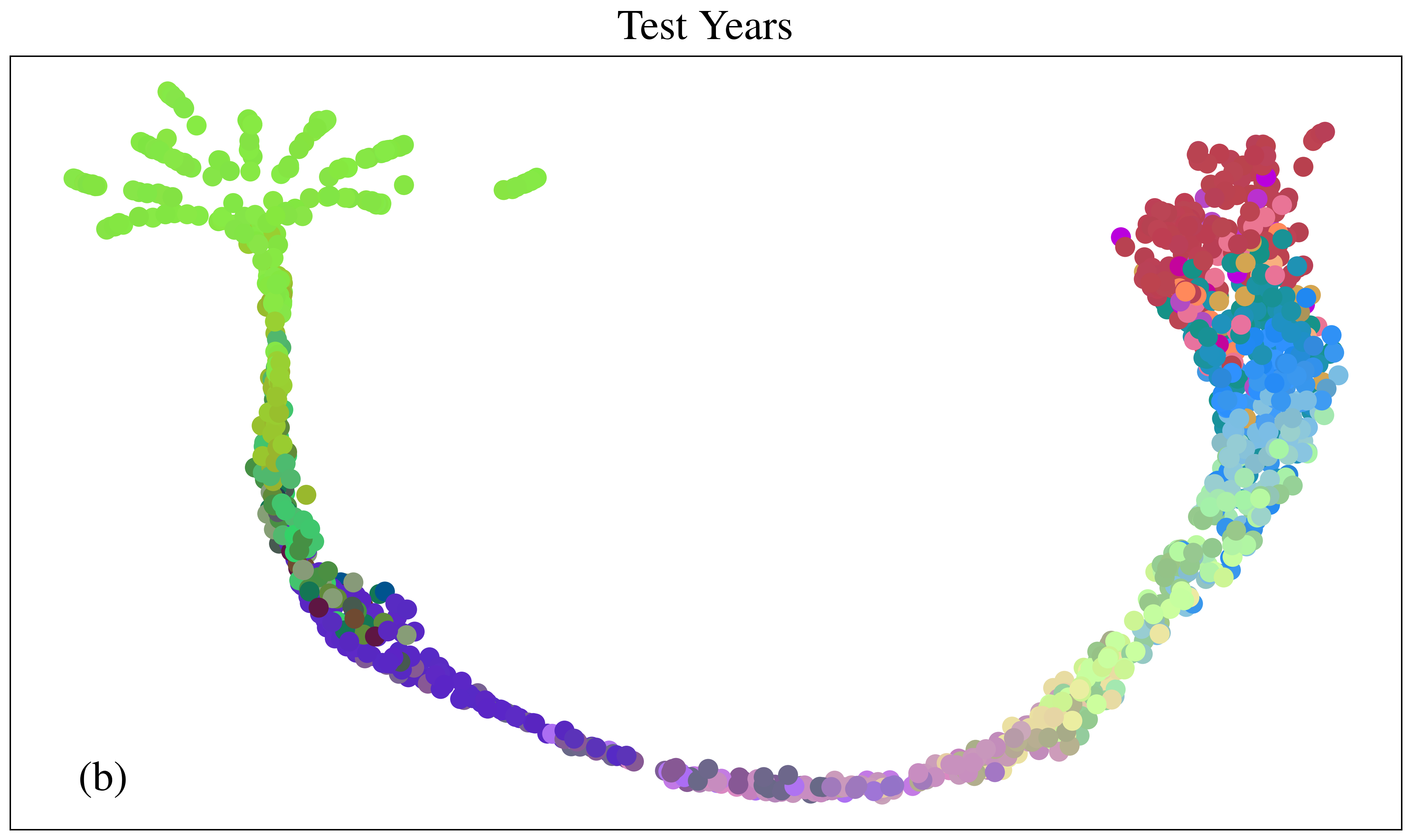}
	\end{subfigure}
	
	\begin{subfigure}{0.45\linewidth}
		\includegraphics[width=\linewidth]{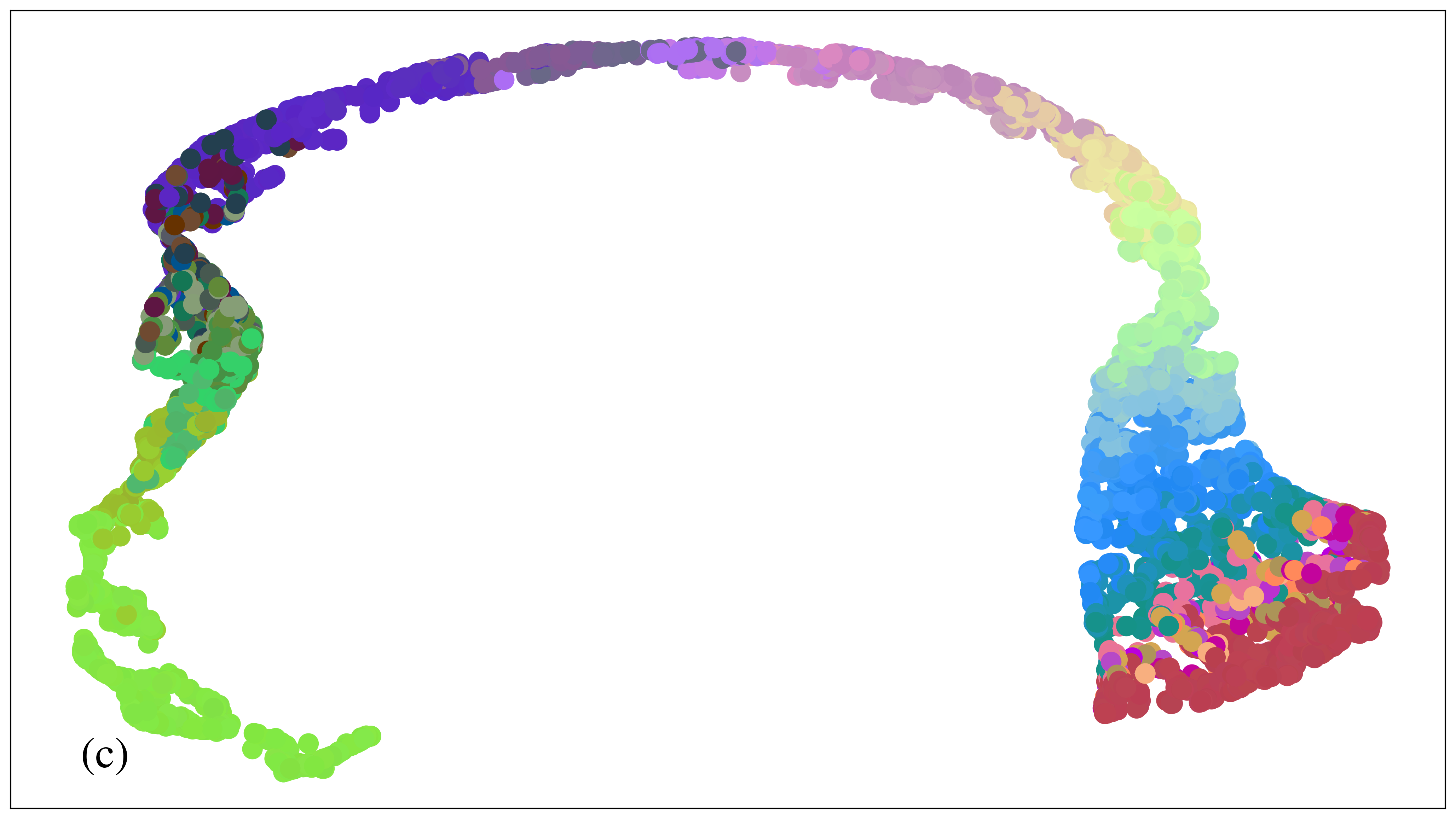}
	\end{subfigure}
	\hfil
	\begin{subfigure}{0.45\linewidth}
		\includegraphics[width=\linewidth]{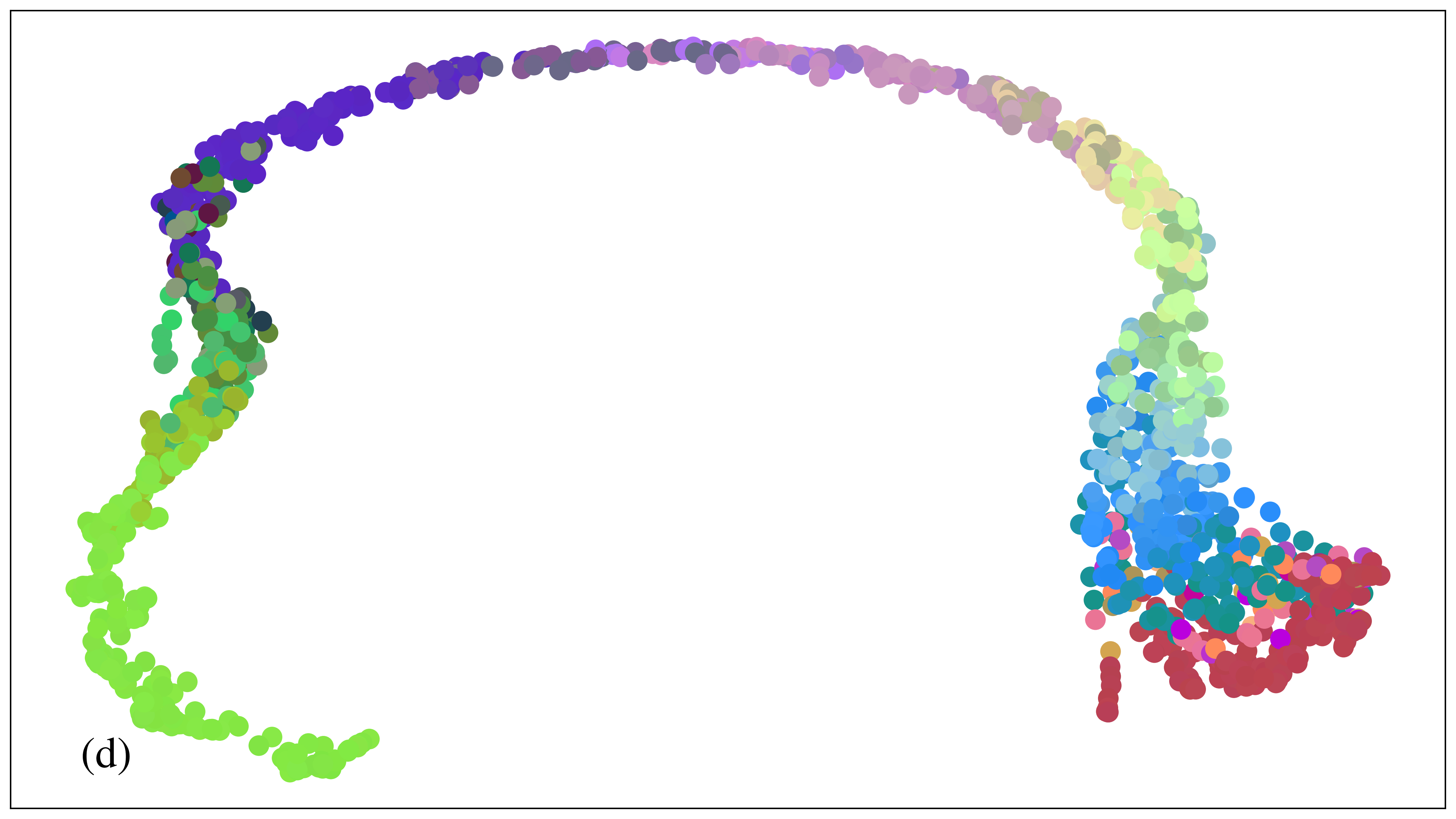}
	\end{subfigure}
	
	\begin{subfigure}{0.45\linewidth}
		\includegraphics[width=\linewidth]{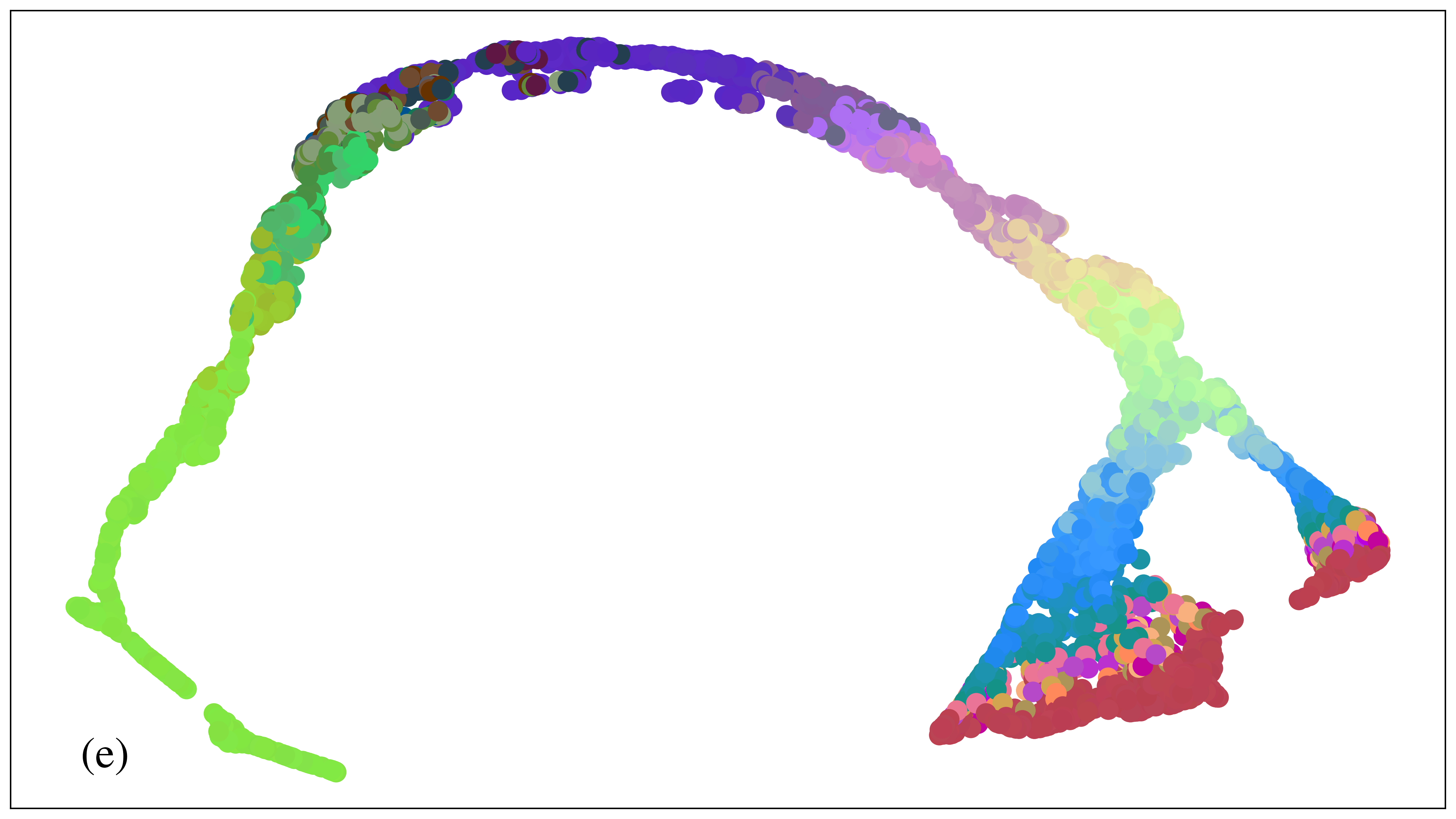}
	\end{subfigure}
	\hfil
	\begin{subfigure}{0.45\linewidth}
		\includegraphics[width=\linewidth]{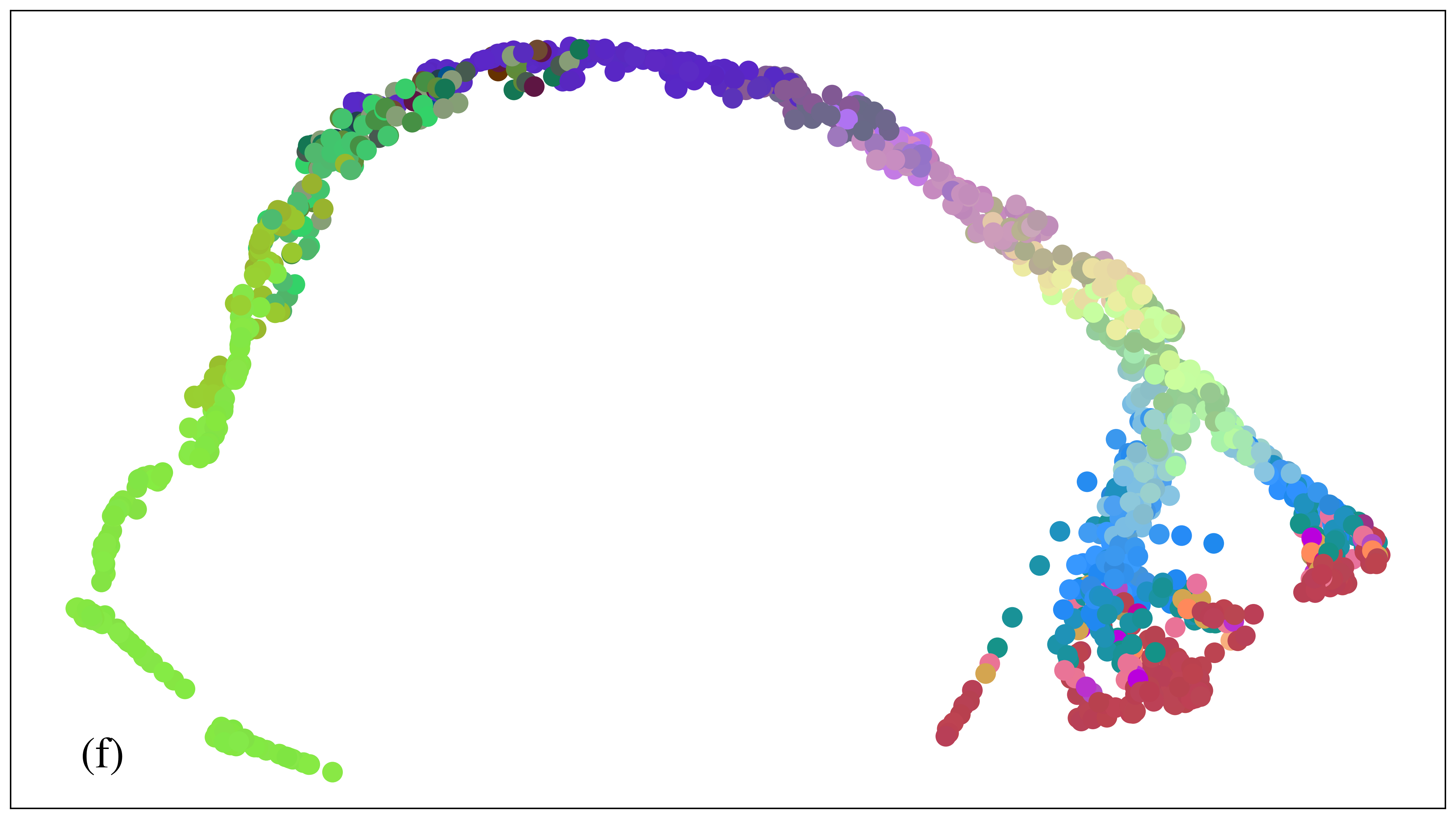}
	\end{subfigure}
	
	\begin{subfigure}{0.9\linewidth}
		\includegraphics[width=\linewidth]{figs/colorbar.png}
	\end{subfigure}

	\caption{UMAP visualizations of 128 node layer activations feeding into the softmax layer for training and test years for (a) and (b) Dense NN; (c) and (d) Sequential NN; (e) and (f) DgNN.}
	\label{fig:UMAP_softmax}
\end{figure}
\begin{paracol}{2}
\linenumbers
\switchcolumn

\section{Conclusions}
In this study, an agronomy-informed neural network, DgNN, was developed to provide in-season CGSE estimates. The DgNN separates inputs that can be treated as independent crop growth drivers using a branched structure, and uses attention mechanisms to account for the varying importance of inputs during the growing season. The DgNN was trained and evaluated on RS and USDA CPR data for Iowa from 2003 to 2019 using NSE and CS as metrics. The DgNN structure was compared to a HMM and two NN structures of similar complexity. The DgNN outperformed each of the other methods on all growth stages for five-fold cross validation on the training data, with an average improvement in NSE across all stages of 22\% versus the HMM and 2.2\% versus the next best NN. The four models were evaluated on four test years that remained unseen during validation. The mean performance during model evaluation was also higher for the DgNN than the other NNs and the HMM. Mean evaluation NSE for the DgNN across all stages was 43\% versus the HMM and 4.0\% higher versus the next best NN (5.9\% when excluding Silking). The DgNN also had 39\% more weeks with the highest CS across all test years than the next best NN. CS metrics showed that weeks when a region's crop is in multiple stages concurrently are more difficult to estimate. Estimating timing of the short yet critical growth stage of Silking was the most difficult for all methods, particularly the Silking-to-Grainfill stage transition. During evaluation, Silking NSE for the DgNN was reduced primarily due to the DgNN's inability to correctly estimate the timing of this transition. While performance of all methods was lower on the test data, LSTM-based methods, the DgNN and Sequential NN, were found to be more robust when presented with abnormal crop progress during model evaluation. UMAP analysis of hidden layers indicated that LSTM-based NNs hold time series from different locations more separate in the activation space.

This study demonstrated that a domain-guided design, such as the DgNN, can improve in-season CGSE compared to NN structures of equivalent complexity. However, UMAP-based NN structure diagnostics and ablation studies to investigate optimum NN complexity may be able to further improve upon these results. In addition, NN boosting methods may also address stage-specific shortcomings, such as Silking-Grainfill transition.

%%%%%%%%%%%%%%%%%%%%%%%%%%%%%%%%%%%%%%%%%%
\vspace{6pt} 

%%%%%%%%%%%%%%%%%%%%%%%%%%%%%%%%%%%%%%%%%%

%%%%%%%%%%%%%%%%%%%%%%%%%%%%%%%%%%%%%%%%%%
\authorcontributions{Conceptualization, all authors; methodology, all authors; validation, G.W.; investigation, all authors; writing---original draft preparation, G.W.; writing---review and editing, all authors; visualization, G.W. All authors have read and agreed to the published version of the manuscript.}

\funding{This research was supported by funding from the NASA Terrestrial Hydrology Program (Grant No. NNX16AQ24G). }

\conflictsofinterest{The authors declare no conflict of interest.}

\end{paracol}

\reftitle{References}

%=====================================
% References, variant A: external bibliography
%=====================================
\externalbibliography{yes}

\bibliography{refs/refs_CGSE.bib, refs/ref_CERESMAIZE.bib}

\begin{thebibliography}{999}

\bibitem[Ray \em{et~al.}(2015)Ray, Gerber, MacDonald, and
  West]{ray_climate_2015}
Ray, D.K.; Gerber, J.S.; MacDonald, G.K.; West, P.C.
\newblock Climate variation explains a third of global crop yield variability.
\newblock {\em Nature Communications} {\bf 2015}, {\em 6},~5989.
\newblock Number: 1,
  doi:{\changeurlcolor{black}\href{https://doi.org/10.1038/ncomms6989}{\detokenize{10.1038/ncomms6989}}}.

\bibitem[Iizumi and Ramankutty(2016)]{iizumi_changes_2016}
Iizumi, T.; Ramankutty, N.
\newblock Changes in yield variability of major crops for 1981–2010 explained
  by climate change.
\newblock {\em Environmental Research Letters} {\bf 2016}, {\em 11},~034003.
\newblock Number: 3,
  doi:{\changeurlcolor{black}\href{https://doi.org/10.1088/1748-9326/11/3/034003}{\detokenize{10.1088/1748-9326/11/3/034003}}}.

\bibitem[Mehrabi and Ramankutty(2019)]{mehrabi_synchronized_2019}
Mehrabi, Z.; Ramankutty, N.
\newblock Synchronized failure of global crop production.
\newblock {\em Nature Ecology \& Evolution} {\bf 2019}, {\em 3},~780--786.
\newblock Number: 5,
  doi:{\changeurlcolor{black}\href{https://doi.org/10.1038/s41559-019-0862-x}{\detokenize{10.1038/s41559-019-0862-x}}}.

\bibitem[Shen \em{et~al.}(2013)Shen, Wu, Di, Yu, Tang, Yu, and
  Shao]{shen_hidden_2013}
Shen, Y.; Wu, L.; Di, L.; Yu, G.; Tang, H.; Yu, G.; Shao, Y.
\newblock Hidden {Markov} {Models} for {Real}-{Time} {Estimation} of {Corn}
  {Progress} {Stages} {Using} {MODIS} and {Meteorological} {Data}.
\newblock {\em Remote Sensing} {\bf 2013}, {\em 5},~1734--1753.
\newblock Number: 4,
  doi:{\changeurlcolor{black}\href{https://doi.org/10.3390/rs5041734}{\detokenize{10.3390/rs5041734}}}.

\bibitem[Zeng \em{et~al.}(2016)Zeng, Wardlow, Wang, Shan, Tadesse, Hayes, and
  Li]{zeng_hybrid_2016}
Zeng, L.; Wardlow, B.D.; Wang, R.; Shan, J.; Tadesse, T.; Hayes, M.J.; Li, D.
\newblock A hybrid approach for detecting corn and soybean phenology with
  time-series {MODIS} data.
\newblock {\em Remote Sensing of Environment} {\bf 2016}, {\em 181},~237--250.
\newblock Publisher: Elsevier Inc.,
  doi:{\changeurlcolor{black}\href{https://doi.org/10.1016/j.rse.2016.03.039}{\detokenize{10.1016/j.rse.2016.03.039}}}.

\bibitem[Seo \em{et~al.}(2019)Seo, Lee, Lee, Hong, and
  Kang]{seo_improving_2019}
Seo, B.; Lee, J.; Lee, K.D.; Hong, S.; Kang, S.
\newblock Improving remotely-sensed crop monitoring by {NDVI}-based crop
  phenology estimators for corn and soybeans in {Iowa} and {Illinois}, {USA}.
\newblock {\em Field Crops Research} {\bf 2019}, p.~16.

\bibitem[Diao(2020)]{diao_remote_2020}
Diao, C.
\newblock Remote sensing phenological monitoring framework to characterize corn
  and soybean physiological growing stages.
\newblock {\em Remote Sensing of Environment} {\bf 2020}, {\em 248}.
\newblock Publisher: Elsevier Inc.,
  doi:{\changeurlcolor{black}\href{https://doi.org/10.1016/j.rse.2020.111960}{\detokenize{10.1016/j.rse.2020.111960}}}.

\bibitem[Ghamghami \em{et~al.}(2020)Ghamghami, Ghahreman, Irannejad, and
  Pezeshk]{ghamghami_parametric_2020}
Ghamghami, M.; Ghahreman, N.; Irannejad, P.; Pezeshk, H.
\newblock A parametric empirical {Bayes} ({PEB}) approach for estimating maize
  progress percentage at field scale.
\newblock {\em Agricultural and Forest Meteorology} {\bf 2020}, {\em
  281},~107829.
\newblock
  doi:{\changeurlcolor{black}\href{https://doi.org/10.1016/j.agrformet.2019.107829}{\detokenize{10.1016/j.agrformet.2019.107829}}}.

\bibitem[Orynbaikyzy \em{et~al.}(2019)Orynbaikyzy, Gessner, and
  Conrad]{orynbaikyzy_crop_2019}
Orynbaikyzy, A.; Gessner, U.; Conrad, C.
\newblock Crop type classification using a combination of optical and radar
  remote sensing data: a review.
\newblock {\em International Journal of Remote Sensing} {\bf 2019}, {\em
  40},~6553--6595.
\newblock Number: 17,
  doi:{\changeurlcolor{black}\href{https://doi.org/10.1080/01431161.2019.1569791}{\detokenize{10.1080/01431161.2019.1569791}}}.

\bibitem[Jia \em{et~al.}(2019)Jia, Khandelwal, Mulla, Pardey, and
  Kumar]{jia_bringing_2019}
Jia, X.; Khandelwal, A.; Mulla, D.J.; Pardey, P.G.; Kumar, V.
\newblock Bringing automated, remote‐sensed, machine learning methods to
  monitoring crop landscapes at scale.
\newblock {\em Agricultural Economics} {\bf 2019}, {\em 50},~41--50.
\newblock
  doi:{\changeurlcolor{black}\href{https://doi.org/10.1111/agec.12531}{\detokenize{10.1111/agec.12531}}}.

\bibitem[Kerner \em{et~al.}(2020)Kerner, Sahajpal, Skakun, Becker-Reshef,
  Barker, Hosseini, Puricelli, and Gray]{kerner_resilient_2020}
Kerner, H.; Sahajpal, R.; Skakun, S.; Becker-Reshef, I.; Barker, B.; Hosseini,
  M.; Puricelli, E.; Gray, P.
\newblock Resilient {In}-{Season} {Crop} {Type} {Classification} in
  {Multispectral} {Satellite} {Observations} using {Growth} {Stage}
  {Normalization}.
\newblock {\em arXiv:2009.10189 [cs, eess]} {\bf 2020}.
\newblock arXiv: 2009.10189.

\bibitem[Teimouri \em{et~al.}(2019)Teimouri, Dyrmann, and
  Jørgensen]{teimouri_novel_2019}
Teimouri, N.; Dyrmann, M.; Jørgensen, R.N.
\newblock A {Novel} {Spatio}-{Temporal} {FCN}-{LSTM} {Network} for
  {Recognizing} {Various} {Crop} {Types} {Using} {Multi}-{Temporal} {Radar}
  {Images}.
\newblock {\em Remote Sensing} {\bf 2019}, {\em 11},~990.
\newblock
  doi:{\changeurlcolor{black}\href{https://doi.org/10.3390/rs11080990}{\detokenize{10.3390/rs11080990}}}.

\bibitem[Weiss \em{et~al.}(2020)Weiss, Jacob, and Duveiller]{Weiss2020}
Weiss, M.; Jacob, F.; Duveiller, G.
\newblock Remote sensing for agricultural applications: {A} meta-review.
\newblock {\em Remote Sensing of Environment} {\bf 2020}, {\em 236},~111402.
\newblock Number: December 2018 Publisher: Elsevier,
  doi:{\changeurlcolor{black}\href{https://doi.org/10.1016/j.rse.2019.111402}{\detokenize{10.1016/j.rse.2019.111402}}}.

\bibitem[Service(2015)]{service_cropscape_2015}
Service, U.N.A.S.
\newblock {CropScape} - {Cropland} {Data} {Layer},  2015.
\newblock type: dataset,
  doi:{\changeurlcolor{black}\href{https://doi.org/10.15482/USDA.ADC/1227096}{\detokenize{10.15482/USDA.ADC/1227096}}}.

\bibitem[Karpatne \em{et~al.}(2017)Karpatne, Atluri, Faghmous, Steinbach,
  Banerjee, Ganguly, Shekhar, Samatova, and Kumar]{karpatne_theory-guided_2017}
Karpatne, A.; Atluri, G.; Faghmous, J.H.; Steinbach, M.; Banerjee, A.; Ganguly,
  A.; Shekhar, S.; Samatova, N.; Kumar, V.
\newblock Theory-{Guided} {Data} {Science}: {A} {New} {Paradigm} for
  {Scientific} {Discovery} from {Data}.
\newblock {\em IEEE Transactions on Knowledge and Data Engineering} {\bf 2017},
  {\em 29},~2318--2331.
\newblock Number: 10,
  doi:{\changeurlcolor{black}\href{https://doi.org/10.1109/TKDE.2017.2720168}{\detokenize{10.1109/TKDE.2017.2720168}}}.

\bibitem[Khandelwal \em{et~al.}(2020)Khandelwal, Xu, Li, Jia, Stienbach, Duffy,
  Nieber, and Kumar]{khandelwal_physics_2020}
Khandelwal, A.; Xu, S.; Li, X.; Jia, X.; Stienbach, M.; Duffy, C.; Nieber, J.;
  Kumar, V.
\newblock Physics {Guided} {Machine} {Learning} {Methods} for {Hydrology}.
\newblock {\em arXiv:2012.02854 [physics]} {\bf 2020}.
\newblock arXiv: 2012.02854.

\bibitem[Willard \em{et~al.}(2020)Willard, Jia, Xu, Steinbach, and
  Kumar]{willard_integrating_2020}
Willard, J.; Jia, X.; Xu, S.; Steinbach, M.; Kumar, V.
\newblock Integrating {Physics}-{Based} {Modeling} with {Machine} {Learning}:
  {A} {Survey}.
\newblock {\em arXiv:2003.04919 [physics, stat]} {\bf 2020}.
\newblock arXiv: 2003.04919.

\bibitem[Karpatne \em{et~al.}(2018)Karpatne, Watkins, Read, and
  Kumar]{karpatne_physics-guided_2018}
Karpatne, A.; Watkins, W.; Read, J.; Kumar, V.
\newblock Physics-guided {Neural} {Networks} ({PGNN}): {An} {Application} in
  {Lake} {Temperature} {Modeling}.
\newblock {\em arXiv:1710.11431 [physics, stat]} {\bf 2018}.
\newblock arXiv: 1710.11431.

\bibitem[Hu \em{et~al.}(2020)Hu, Hu, Verma, and Zhang]{hu_physics-guided_2020}
Hu, X.; Hu, H.; Verma, S.; Zhang, Z.L.
\newblock Physics-{Guided} {Deep} {Neural} {Networks} for {Power} {Flow}
  {Analysis}.
\newblock {\em IEEE Transactions on Power Systems} {\bf 2020}, pp. 1--1.
\newblock
  doi:{\changeurlcolor{black}\href{https://doi.org/10.1109/TPWRS.2020.3029557}{\detokenize{10.1109/TPWRS.2020.3029557}}}.

\bibitem[Rong \em{et~al.}(2020)Rong, Zhang, and Wang]{rong_lagrangian_2020}
Rong, M.; Zhang, D.; Wang, N.
\newblock A {Lagrangian} {Dual}-based {Theory}-guided {Deep} {Neural}
  {Network}.
\newblock {\em arXiv:2008.10159 [cs, math, stat]} {\bf 2020}.
\newblock arXiv: 2008.10159.

\bibitem[Wang \em{et~al.}(2020)Wang, Zhang, Chang, and Li]{wang_deep_2020}
Wang, N.; Zhang, D.; Chang, H.; Li, H.
\newblock Deep learning of subsurface flow via theory-guided neural network.
\newblock {\em Journal of Hydrology} {\bf 2020}, {\em 584},~124700.
\newblock
  doi:{\changeurlcolor{black}\href{https://doi.org/10.1016/j.jhydrol.2020.124700}{\detokenize{10.1016/j.jhydrol.2020.124700}}}.

\bibitem[Service(2021)]{usda_foreign_agricultural_service_world_2021}
Service, U.F.A.
\newblock World {Agricultural} {Production}.
\newblock Technical Report WAP 4-21, USDA Foreign Agricultural Service,  2021.

\bibitem[{USDA National Agricultural Statistics
  Service}()]{usda_national_agricultural_statistics_service_corn_nodate}
{USDA National Agricultural Statistics Service}.
\newblock Corn - {Acres} {Planted}.

\bibitem[Wang \em{et~al.}(2020)Wang, Di~Tommaso, Deines, and
  Lobell]{wang_mapping_2020}
Wang, S.; Di~Tommaso, S.; Deines, J.M.; Lobell, D.B.
\newblock Mapping twenty years of corn and soybean across the {US} {Midwest}
  using the {Landsat} archive.
\newblock {\em Scientific Data} {\bf 2020}, {\em 7},~307.
\newblock
  doi:{\changeurlcolor{black}\href{https://doi.org/10.1038/s41597-020-00646-4}{\detokenize{10.1038/s41597-020-00646-4}}}.

\bibitem[{USDA Ag Data Commons}()]{usda_ag_data_commons_cropscape_nodate}
{USDA Ag Data Commons}.
\newblock {CropScape} - {Cropland} {Data} {Layer}.

\bibitem[Lizaso \em{et~al.}(2011)Lizaso, Boote, Jones, Porter, Echarte,
  Westgate, and Sonohat]{lizaso_csm-ixim_2011}
Lizaso, J.I.; Boote, K.J.; Jones, J.W.; Porter, C.H.; Echarte, L.; Westgate,
  M.E.; Sonohat, G.
\newblock {CSM}-{IXIM}: {A} {New} {Maize} {Simulation} {Model} for {DSSAT}
  {Version} 4.5.
\newblock {\em Agronomy Journal} {\bf 2011}, {\em 103},~766--779.
\newblock
  doi:{\changeurlcolor{black}\href{https://doi.org/10.2134/agronj2010.0423}{\detokenize{10.2134/agronj2010.0423}}}.

\bibitem[Kiniry(2015)]{hanks_maize_2015}
Kiniry, J.R.
\newblock Maize {Phasic} {Development}. In {\em Agronomy {Monographs}}; Hanks,
  J.; Ritchie, J.T., Eds.; American Society of Agronomy, Crop Science Society
  of America, Soil Science Society of America: Madison, WI, USA,  2015; pp.
  55--70.
\newblock
  doi:{\changeurlcolor{black}\href{https://doi.org/10.2134/agronmonogr31.c4}{\detokenize{10.2134/agronmonogr31.c4}}}.

\bibitem[Thornton \em{et~al.}(2020)Thornton, Shrestha, Wei, Thornton, Kao, and
  Wilson]{thornton_daymet_2020}
Thornton, M.; Shrestha, R.; Wei, Y.; Thornton, P.; Kao, S.; Wilson, B.
\newblock {\em Daymet: {Daily} {Surface} {Weather} {Data} on a 1-km {Grid} for
  {North} {America}, {Version} 4}; ORNL Distributed Active Archive Center,
  2020.
\newblock
  doi:{\changeurlcolor{black}\href{https://doi.org/10.3334/ORNLDAAC/1840}{\detokenize{10.3334/ORNLDAAC/1840}}}.

\bibitem[{R. Myneni, Y. Knyazikhin, T.
  Park.}(2015)]{r_myneni_y_knyazikhin_t_park_mcd15a3h_2015}
{R. Myneni, Y. Knyazikhin, T. Park.}.
\newblock {MCD15A3H} {MODIS}/{Terra}+{Aqua} {Leaf} {Area} {Index}/{FPAR} 4-day
  {L4} {Global} 500m {SIN} {Grid} {V006}.
\newblock {\em NASA EOSDIS Land Processes DAAC} {\bf 2015}.
\newblock
  doi:{\changeurlcolor{black}\href{https://doi.org/10.5067/MODIS/MCD15A3H.006}{\detokenize{10.5067/MODIS/MCD15A3H.006}}}.

\bibitem[Staff(2020)]{soil_survey_staff_gridded_2020}
Staff, S.S.
\newblock Gridded {Soil} {Survey} {Geographic} ({gSSURGO}) {Database} for the
  {United} {States} of {America} and the {Territories}, {Commonwealths}, and
  {Island} {Nations} served by the {USDA}-{NRCS}.
\newblock Technical report, USDA National Resource Conversvation Service,
  2020.
\newblock Publisher: United States Department of Agriculture, Natural Resources
  Conservation Service.

\bibitem[Chen \em{et~al.}(2004)Chen, Jönsson, Tamura, Gu, Matsushita, and
  Eklundh]{chen_simple_2004}
Chen, J.; Jönsson, P.; Tamura, M.; Gu, Z.; Matsushita, B.; Eklundh, L.
\newblock A simple method for reconstructing a high-quality {NDVI} time-series
  data set based on the {Savitzky}–{Golay} filter.
\newblock {\em Remote Sensing of Environment} {\bf 2004}, {\em 91},~332--344.
\newblock Number: 3-4,
  doi:{\changeurlcolor{black}\href{https://doi.org/10.1016/j.rse.2004.03.014}{\detokenize{10.1016/j.rse.2004.03.014}}}.

\bibitem[Hochreiter and Schmidhuber(1997)]{hochreiter_long_1997}
Hochreiter, S.; Schmidhuber, J.
\newblock Long {Short}-{Term} {Memory}.
\newblock {\em Neural Computation} {\bf 1997}, {\em 9},~1735--1780.
\newblock Number: 8,
  doi:{\changeurlcolor{black}\href{https://doi.org/10.1162/neco.1997.9.8.1735}{\detokenize{10.1162/neco.1997.9.8.1735}}}.

\bibitem[Yu \em{et~al.}(2019)Yu, Si, Hu, and Zhang]{yu_review_2019}
Yu, Y.; Si, X.; Hu, C.; Zhang, J.
\newblock A {Review} of {Recurrent} {Neural} {Networks}: {LSTM} {Cells} and
  {Network} {Architectures}.
\newblock {\em Neural Computation} {\bf 2019}, {\em 31},~1235--1270.
\newblock
  doi:{\changeurlcolor{black}\href{https://doi.org/10.1162/neco_a_01199}{\detokenize{10.1162/neco_a_01199}}}.

\bibitem[Khaki \em{et~al.}(2020)Khaki, Wang, and
  Archontoulis]{khaki_cnn-rnn_2020}
Khaki, S.; Wang, L.; Archontoulis, S.V.
\newblock A {CNN}-{RNN} {Framework} for {Crop} {Yield} {Prediction}.
\newblock {\em Frontiers in Plant Science} {\bf 2020}, {\em 10},~1750.
\newblock
  doi:{\changeurlcolor{black}\href{https://doi.org/10.3389/fpls.2019.01750}{\detokenize{10.3389/fpls.2019.01750}}}.

\bibitem[Graves(2012)]{graves_long_2012}
Graves, A.
\newblock Long short-term memory. In {\em Supervised sequence labelling with
  recurrent neural networks}; Springer,  2012; pp. 37--45.

\bibitem[Warrington and Kanemasu(1983{\natexlab{a}})]{warrington_corn_1983}
Warrington, I.J.; Kanemasu, E.T.
\newblock Corn {Growth} {Response} to {Temperature} and {Photoperiod} {I}.
  {Seedling} {Emergence}, {Tassel} {Initiation}, and {Anthesis}
  $^{\textrm{1}}$.
\newblock {\em Agronomy Journal} {\bf 1983}, {\em 75},~749--754.
\newblock
  doi:{\changeurlcolor{black}\href{https://doi.org/10.2134/agronj1983.00021962007500050008x}{\detokenize{10.2134/agronj1983.00021962007500050008x}}}.

\bibitem[Warrington and Kanemasu(1983{\natexlab{b}})]{warrington_corn_1983-1}
Warrington, I.J.; Kanemasu, E.T.
\newblock Corn {Growth} {Response} to {Temperature} and {Photoperiod} {II}.
  {Leaf}‐{Initiation} and {Leaf}‐{Appearance} {Rates} $^{\textrm{1}}$.
\newblock {\em Agronomy Journal} {\bf 1983}, {\em 75},~755--761.
\newblock
  doi:{\changeurlcolor{black}\href{https://doi.org/10.2134/agronj1983.00021962007500050009x}{\detokenize{10.2134/agronj1983.00021962007500050009x}}}.

\bibitem[NeSmith and Ritchie(1992)]{nesmith_short_1992}
NeSmith, D.; Ritchie, J.
\newblock Short‐ and {Long}‐{Term} {Responses} of {Corn} to a
  {Pre}‐{Anthesis} {Soil} {Water} {Deficit}.
\newblock {\em Agronomy Journal} {\bf 1992}, {\em 84},~107--113.
\newblock
  doi:{\changeurlcolor{black}\href{https://doi.org/10.2134/agronj1992.00021962008400010021x}{\detokenize{10.2134/agronj1992.00021962008400010021x}}}.

\bibitem[Çakir(2004)]{cakir_effect_2004}
Çakir, R.
\newblock Effect of water stress at different development stages on vegetative
  and reproductive growth of corn.
\newblock {\em Field Crops Research} {\bf 2004}, {\em 89},~1--16.
\newblock
  doi:{\changeurlcolor{black}\href{https://doi.org/10.1016/j.fcr.2004.01.005}{\detokenize{10.1016/j.fcr.2004.01.005}}}.

\bibitem[Jones \em{et~al.}(1983)Jones, Ritchie, Kiniry, Godwin, and
  Otter]{jones1983ceres}
Jones, C.; Ritchie, J.; Kiniry, J.; Godwin, D.; Otter, S.
\newblock The CERES wheat and maize models.
\newblock  Proceedings of the International Symposium on Minimum Data Sets for
  Agrotechnology Transfer, ICRISAT Center, India,  1983, pp. 95--100.

\bibitem[Holzworth \em{et~al.}(2014)Holzworth, Huth, deVoil, Zurcher, Herrmann,
  McLean, Chenu, van Oosterom, Snow, Murphy, Moore, Brown, Whish, Verrall,
  Fainges, Bell, Peake, Poulton, Hochman, Thorburn, Gaydon, Dalgliesh,
  Rodriguez, Cox, Chapman, Doherty, Teixeira, Sharp, Cichota, Vogeler, Li,
  Wang, Hammer, Robertson, Dimes, Whitbread, Hunt, van Rees, McClelland,
  Carberry, Hargreaves, MacLeod, McDonald, Harsdorf, Wedgwood, and
  Keating]{holzworth_apsim_2014}
Holzworth, D.P.; Huth, N.I.; deVoil, P.G.; Zurcher, E.J.; Herrmann, N.I.;
  McLean, G.; Chenu, K.; van Oosterom, E.J.; Snow, V.; Murphy, C.; Moore, A.D.;
  Brown, H.; Whish, J.P.; Verrall, S.; Fainges, J.; Bell, L.W.; Peake, A.S.;
  Poulton, P.L.; Hochman, Z.; Thorburn, P.J.; Gaydon, D.S.; Dalgliesh, N.P.;
  Rodriguez, D.; Cox, H.; Chapman, S.; Doherty, A.; Teixeira, E.; Sharp, J.;
  Cichota, R.; Vogeler, I.; Li, F.Y.; Wang, E.; Hammer, G.L.; Robertson, M.J.;
  Dimes, J.P.; Whitbread, A.M.; Hunt, J.; van Rees, H.; McClelland, T.;
  Carberry, P.S.; Hargreaves, J.N.; MacLeod, N.; McDonald, C.; Harsdorf, J.;
  Wedgwood, S.; Keating, B.A.
\newblock {APSIM} – {Evolution} towards a new generation of agricultural
  systems simulation.
\newblock {\em Environmental Modelling \& Software} {\bf 2014}, {\em
  62},~327--350.
\newblock
  doi:{\changeurlcolor{black}\href{https://doi.org/10.1016/j.envsoft.2014.07.009}{\detokenize{10.1016/j.envsoft.2014.07.009}}}.

\bibitem[Young \em{et~al.}(2018)Young, Hazarika, Poria, and
  Cambria]{young_recent_2018}
Young, T.; Hazarika, D.; Poria, S.; Cambria, E.
\newblock Recent {Trends} in {Deep} {Learning} {Based} {Natural} {Language}
  {Processing} [{Review} {Article}].
\newblock {\em IEEE Computational Intelligence Magazine} {\bf 2018}, {\em
  13},~55--75.
\newblock
  doi:{\changeurlcolor{black}\href{https://doi.org/10.1109/MCI.2018.2840738}{\detokenize{10.1109/MCI.2018.2840738}}}.

\bibitem[Liu \em{et~al.}(2016)Liu, Sun, Lin, and Wang]{liu_learning_2016}
Liu, Y.; Sun, C.; Lin, L.; Wang, X.
\newblock Learning {Natural} {Language} {Inference} using {Bidirectional}
  {LSTM} model and {Inner}-{Attention}.
\newblock {\em arXiv:1605.09090 [cs]} {\bf 2016}.
\newblock arXiv: 1605.09090.

\bibitem[Vaswani \em{et~al.}(2017)Vaswani, Shazeer, Parmar, Uszkoreit, Jones,
  Gomez, Kaiser, and Polosukhin]{vaswani_attention_2017}
Vaswani, A.; Shazeer, N.; Parmar, N.; Uszkoreit, J.; Jones, L.; Gomez, A.N.;
  Kaiser, L.; Polosukhin, I.
\newblock Attention {Is} {All} {You} {Need}.
\newblock {\em arXiv:1706.03762 [cs]} {\bf 2017}.
\newblock arXiv: 1706.03762.

\bibitem[McInnes \em{et~al.}(2020)McInnes, Healy, and
  Melville]{mcinnes_umap_2020}
McInnes, L.; Healy, J.; Melville, J.
\newblock {UMAP}: {Uniform} {Manifold} {Approximation} and {Projection} for
  {Dimension} {Reduction}.
\newblock {\em arXiv:1802.03426 [cs, stat]} {\bf 2020}.
\newblock arXiv: 1802.03426.

\bibitem[Becht \em{et~al.}(2019)Becht, McInnes, Healy, Dutertre, Kwok, Ng,
  Ginhoux, and Newell]{becht_dimensionality_2019}
Becht, E.; McInnes, L.; Healy, J.; Dutertre, C.A.; Kwok, I.W.H.; Ng, L.G.;
  Ginhoux, F.; Newell, E.W.
\newblock Dimensionality reduction for visualizing single-cell data using
  {UMAP}.
\newblock {\em Nature Biotechnology} {\bf 2019}, {\em 37},~38--44.
\newblock
  doi:{\changeurlcolor{black}\href{https://doi.org/10.1038/nbt.4314}{\detokenize{10.1038/nbt.4314}}}.

\bibitem[{USDA-National Agricultural Statistics Service Upper Midwest Region,
  Iowa Field
  Office}(2010)]{usda-national_agricultural_statistics_service_upper_midwest_region_iowa_field_office_2010_2010}
{USDA-National Agricultural Statistics Service Upper Midwest Region, Iowa Field
  Office}.
\newblock 2010 {Iowa} {Agricultural} {Statistics}.
\newblock Technical report, USDA,  2010.

\bibitem[{USDA-National Agricultural Statistics Service Upper Midwest Region,
  Iowa Field
  Office}(2013)]{usda-national_agricultural_statistics_service_upper_midwest_region_iowa_field_office_2013_2013}
{USDA-National Agricultural Statistics Service Upper Midwest Region, Iowa Field
  Office}.
\newblock 2013 {Iowa} {Agricultural} {Statistics}.
\newblock Technical report, USDA,  2013.

\bibitem[{USDA-National Agricultural Statistics Service Upper Midwest Region,
  Iowa Field
  Office}(2015)]{usda-national_agricultural_statistics_service_upper_midwest_region_iowa_field_office_2015_2015}
{USDA-National Agricultural Statistics Service Upper Midwest Region, Iowa Field
  Office}.
\newblock 2015 {Iowa} {Agricultural} {Statistics}.
\newblock Technical report, USDA,  2015.

\bibitem[{USDA-National Agricultural Statistics Service Upper Midwest Region,
  Iowa Field
  Office}(2020)]{usda-national_agricultural_statistics_service_upper_midwest_region_iowa_field_office_2020_2020}
{USDA-National Agricultural Statistics Service Upper Midwest Region, Iowa Field
  Office}.
\newblock 2020 {Iowa} {Agricultural} {Statistics}.
\newblock Technical report, USDA,  2020.

\bibitem[{USDA National Agricultural Statistics
  Service}()]{usda_national_agricultural_statistics_service_corn_nodate-1}
{USDA National Agricultural Statistics Service}.
\newblock Corn - {Crop} {Condition}.

\bibitem[Meyes \em{et~al.}(2019)Meyes, Lu, de~Puiseau, and
  Meisen]{meyes_ablation_2019}
Meyes, R.; Lu, M.; de~Puiseau, C.W.; Meisen, T.
\newblock Ablation {Studies} in {Artificial} {Neural} {Networks}.
\newblock {\em arXiv:1901.08644 [cs, q-bio]} {\bf 2019}.
\newblock arXiv: 1901.08644.

\bibitem[Peerlinck \em{et~al.}(2019)Peerlinck, Sheppard, and
  Senecal]{peerlinck_adaboost_2019}
Peerlinck, A.; Sheppard, J.; Senecal, J.
\newblock {AdaBoost} with {Neural} {Networks} for {Yield} and {Protein}
  {Prediction} in {Precision} {Agriculture}.
\newblock  2019 {International} {Joint} {Conference} on {Neural} {Networks}
  ({IJCNN}); IEEE: Budapest, Hungary,  2019; pp. 1--8.
\newblock
  doi:{\changeurlcolor{black}\href{https://doi.org/10.1109/IJCNN.2019.8851976}{\detokenize{10.1109/IJCNN.2019.8851976}}}.

\end{thebibliography}

% If authors have biography, please use the format below
\section*{Short Biography of Authors}
\bio
{\raisebox{-0.35cm}{\includegraphics[width=3.5cm,height=5.3cm,clip,keepaspectratio]{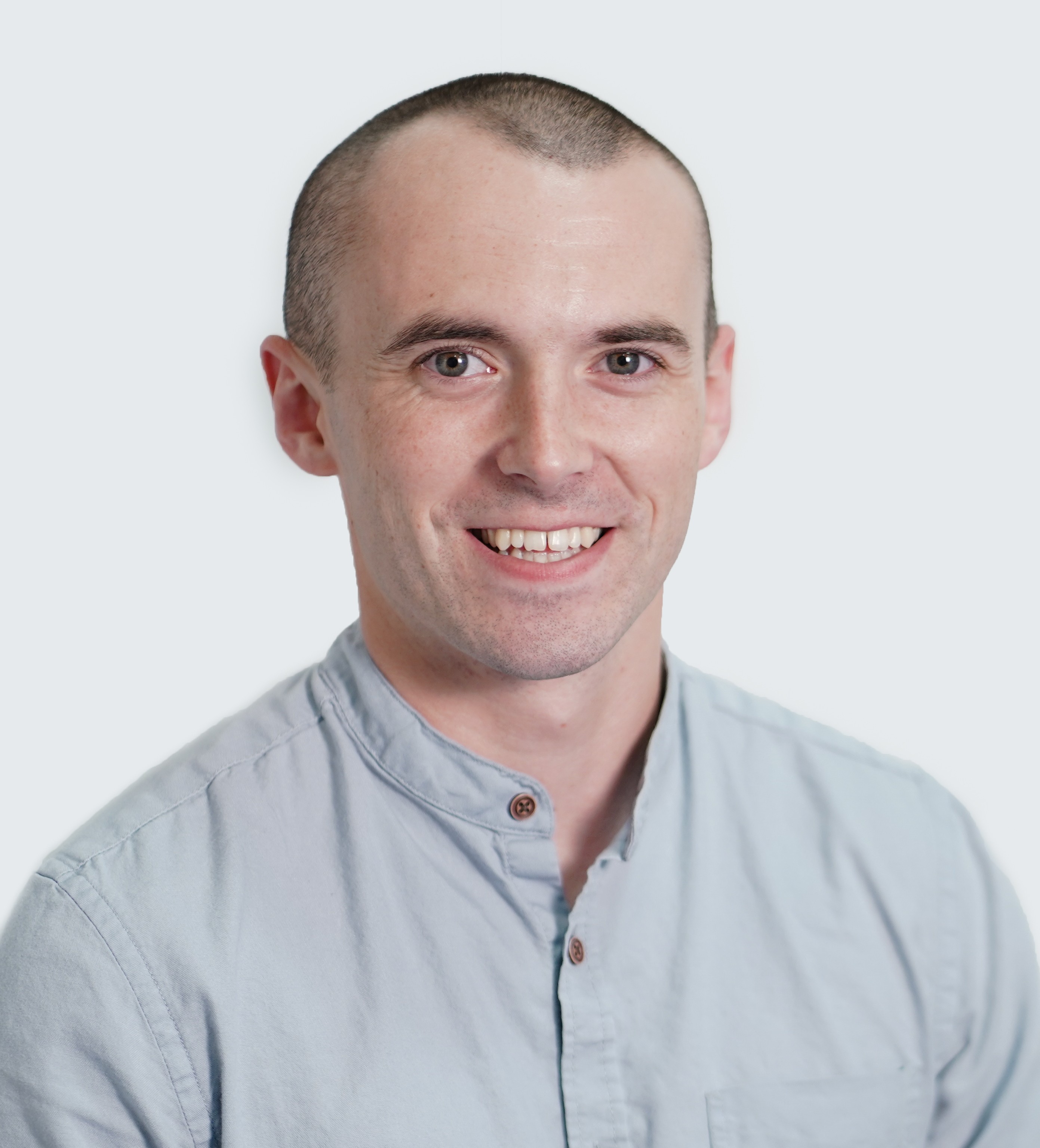}}}
{\textbf{George Worrall} is currently pursuing the Ph.D degree at the Agricultural and Biological Engineering Department, University of Florida, Gainesville, Fl, USA. He received the BEng degree in General Engineering from Durham University in 2017, and the MSc degree in Digital Signal Processing from the University of Manchester in 2018. His research interests include applications of remote sensing in agriculture, physics-guided machine learning, and mechanistic crop modeling.}

\bio
{\raisebox{-0.35cm}{\includegraphics[width=3.5cm,height=5.3cm,clip,keepaspectratio]{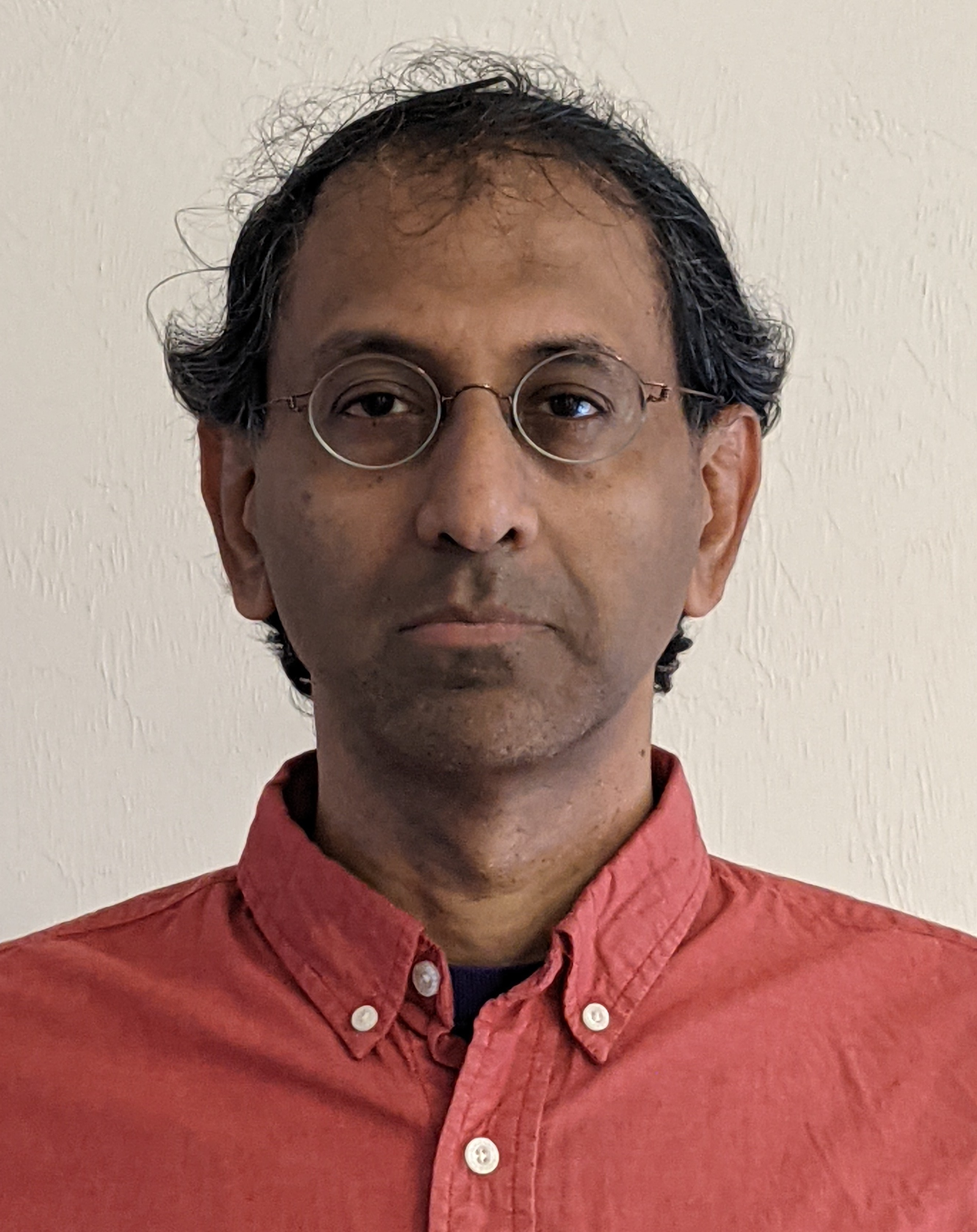}}}
{\textbf{Anand Rangarajan} is currently a Professor with the Department of Computer and Information Science and Engineering, University of Florida, Gainesville, FL, USA. His research interests include computer vision, machine learning, medical and hyperspectral imaging, and the science of consciousness.}

\bio
{\raisebox{-0.35cm}{\includegraphics[width=3.5cm,height=5.3cm,clip,keepaspectratio]{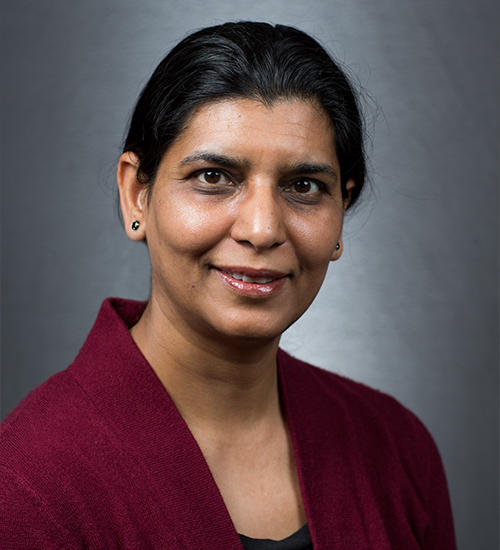}}}
{\textbf{Jasmeet Judge} received the Ph.D. degree in electrical engineering and atmospheric, oceanic, and space sciences from the University of Michigan, Ann Arbor, MI, USA, in 1999.
	She is currently the Director of the Center for Remote Sensing and Professor with the Agricultural and Biological Engineering Department, Institute of Food and Agricultural Sciences, University of Florida, Gainesville, FL, USA. Her research interests include remote sensing applications for soil and crop dynamics in agricultural terrains; spatio-temporal scaling; and data assimilation.}

%%%%%%%%%%%%%%%%%%%%%%%%%%%%%%%%%%%%%%%%%%
%% for journal Sci
%\reviewreports{\\
%Reviewer 1 comments and authors’ response\\
%Reviewer 2 comments and authors’ response\\
%Reviewer 3 comments and authors’ response
%}
%%%%%%%%%%%%%%%%%%%%%%%%%%%%%%%%%%%%%%%%%%
\end{document}